\documentclass[11pt, a4paper, logo, copyright, nonumbering]{map}
\usepackage{CJKutf8}

\usepackage{xargs}  

\usepackage{todonotes}
\usepackage{amsmath}
\usepackage{dsfont}

\usepackage{longtable}

\usepackage{svg}
\usepackage{fontawesome}
\usepackage{array}
\usepackage{tabularx}
\usepackage{latexsym}
\usepackage{graphicx}
\usepackage[utf8]{inputenc} 
\usepackage[utf8]{inputenc} 
\usepackage{amssymb} 
\usepackage{url}            
\usepackage{amsfonts}       
\usepackage{nicefrac}       
\usepackage{microtype}      
\usepackage{xspace}

\usepackage{listings}
\usepackage{makecell}
\usepackage{inconsolata}
\usepackage{multicol}
\usepackage{amstext}

\definecolor{darkblue}{RGB}{84, 112, 198}

\definecolor{lightblue}{rgb}{0.85, 0.95, 1.0}    
\definecolor{lightgreen}{rgb}{0.90, 1.0, 0.90}    
\definecolor{lightorange}{rgb}{1.0, 0.95, 0.85}   
\definecolor{lightpurple}{rgb}{0.95, 0.90, 1.0}   
\definecolor{lightgray}{rgb}{0.97, 0.97, 0.97}    

\usepackage{tikz}
\usetikzlibrary{calc}
\definecolor{battery-empty}{rgb}{0.9, 0.9, 0.9}
\newcommand{\difficultybar}[1]{%
  \begin{tikzpicture}[baseline, scale=0.5, every node/.style={scale=0.8}]
    \foreach \i in {1,2,3,4,5} {
      \ifnum\i>#1
        \draw[fill=battery-empty] (\i*0.5-0.5, 0) rectangle (\i*0.5, 0.25);
      \else
        \pgfmathsetmacro{\colorlevel}{80 - 12*(\i)} 
        \edef\x{\noexpand\draw[fill=blue!\colorlevel!white, opacity=0.9] (\i*0.5-0.5, 0) rectangle (\i*0.5, 0.25);}
        \x
        \draw[blue!50!black] (\i*0.5-0.5, 0) rectangle (\i*0.5, 0.25);
      \fi
    }
    \fill[battery-empty!70] (2.5, 0.08) rectangle (2.6, 0.17);
    \draw[battery-empty!70!black] (2.5, 0.08) rectangle (2.6, 0.17);
  \end{tikzpicture}%
}
\renewcommand{\arraystretch}{0.96}

\definecolor{hidden-draw}{RGB}{20,68,106}
\definecolor{hidden-pink}{RGB}{255,245,247}
\usepackage[edges]{forest}

\usepackage{placeins}
\usepackage{dblfloatfix}

\usepackage{CJKutf8}
\usepackage{xargs}
\usepackage{multirow}
\usepackage{cleveref}
\usepackage{subcaption}
\usepackage{url}
\usepackage{svg}
\usepackage{fontawesome}
\usepackage{xcolor}
\usepackage{float}

\usepackage[utf8]{inputenc}
\usepackage{booktabs}
\usepackage{graphicx}
\usepackage{array}
\usepackage{tabularx}
\usepackage{xcolor,colortbl}  
\definecolor{boxcolor}{HTML}{d92523} 
\definecolor{bulbcolor}{HTML}{e3b87f} 
\usepackage{url}
\usepackage{ragged2e}   
\usepackage{makecell} 

\usepackage{hyperref}       
\usepackage{url}            
\usepackage{booktabs}       
\usepackage{nicefrac}       
\usepackage{microtype}      
\usepackage{xcolor}         
\usepackage{graphicx}
\usepackage{longtable}
\usepackage{array}
\usepackage{pbox}
\usepackage{tabularx}
\usepackage{multirow}
\usepackage{wrapfig}
\usepackage{pifont}
\usepackage{algorithm}
\usepackage{algorithmic}
\usepackage{lipsum}
\usepackage{subcaption}
\usepackage{enumitem}
\usepackage{tikz}
\usepackage{xstring}

\usepackage{color-edits}

\usepackage{booktabs}
\usepackage{multirow}
\usepackage[table]{xcolor}
\usepackage{tabularx}

\usepackage{parskip} 

\usepackage{threeparttable} 

\definecolor{rliableolive}{HTML}{BBCC33}
\definecolor{rliableblue}{HTML}{77AADD}
\definecolor{rliablered}{HTML}{f63c44}

\definecolor{rliableolive}{HTML}{BBCC33}
\definecolor{rliableblue}{HTML}{77AADD}
\definecolor{rliablered}{HTML}{f63c44}

\usepackage[most,skins,theorems]{tcolorbox}
\tcbuselibrary{skins,breakable,fitting,hooks}  
\tcbset{
  aibox/.style={
    width=\linewidth,
    top=7pt,
    bottom=2pt,
    colback=rliablered!18!white,
    colframe=black,
    colbacktitle=black,
    enhanced,
    center,
    attach boxed title to top left={yshift=-0.1in,xshift=0.15in},
    boxed title style={boxrule=0pt,colframe=white,},
  }
}
\tcbset{
  aibox2/.style={
    width=\linewidth,
    top=7pt,
    bottom=2pt,
    colback=green!18!white,
    colframe=black,
    colbacktitle=black,
    enhanced,
    center,
    attach boxed title to top left={yshift=-0.1in,xshift=0.15in},
    boxed title style={boxrule=0pt,colframe=white,},
  }
}
\newtcolorbox{AIbox}[2][]{aibox,title=#2,#1}
\newtcolorbox{AIbox2}[2][]{aibox2,title=#2,#1}

\hypersetup{
    colorlinks=true,            
    linkcolor=blue,             
    filecolor=magenta,          
    urlcolor=cyan,              
    citecolor=purple,             
    pdftitle={X-gram: Efficient Token-Indexed Memory Injection for Transformer Language Models},
    pdfpagemode=FullScreen,
}

\definecolor{iquestblue}{HTML}{173C7F}
\definecolor{iquestazure}{HTML}{528FCC}

\renewcommand{\today}{}
\newcommand{\aname}{\textsc{X-gram}\xspace}
\newcommandx{\info}[2][1=]{\todo[linecolor=red,backgroundcolor=red!25,bordercolor=red,#1]{#2}}
\title{
    \vspace{-0.2in}
    \centering \fontsize{15pt}{16pt}\selectfont
    \raisebox{-0.02\height}{
    } \\  
    Beyond N-gram: Data-Aware \aname Extraction for  \\  Efficient Embedding Parameter Scaling
    \vspace{-0.2in}
}

\author{
\textbf{Yilong Chen$^{1,2,4,*}$, Yanxi Xie$^{3,*}$, Zitian Gao$^{4}$, He Xin$^{4}$, Yihao Xiao$^{4}$, Jason Klein Liu$^{4}$,Haoming Luo$^{4}$, Yifan Luo$^{4}$, Zhengmao Ye$^{4}$, Tingwen Liu$^{1,2}$, Xin Zhao$^{4}$, Ran Tao$^{4,\dagger}$, Bryan Dai$^{4,\dagger}$}

\normalsize $^1$ Institute of Information Engineering, Chinese Academy of Sciences 
\normalsize $^2$ School of Cyber Security, University of Chinese Academy of Sciences 
\normalsize $^3$ Peking University
\normalsize $^4$ IQuest Research \\

\textbf{Code:} \url{https://github.com/Longyichen/X-gram} \\

}

\begin{abstract}
  Large token-indexed lookup tables provide a compute-decoupled scaling path, but their practical gains are often limited by poor parameter efficiency and rapid memory growth. We attribute these limitations to Zipfian under-training of the long tail, heterogeneous demand across layers, and ``slot collapse'' that produces redundant embeddings. To address this, we propose \aname, a \textbf{frequency-aware} \textbf{dynamic} \textbf{token-injection framework}. \aname employs hybrid hashing and alias mixing to compress the tail while preserving head capacity, and refines retrieved vectors via normalized SwiGLU ShortConv to \textbf{extract diverse local $n$-gram features}. These signals are integrated into attention value streams and inter-layer residuals using depth-aware gating, effectively aligning static memory with dynamic context. This design introduces a memory-centric scaling axis that decouples model capacity from FLOPs. Extensive evaluations at the 0.73B and 1.15B scales show that \aname improves average accuracy by as much as \textbf{4.4 points} over the vanilla backbone and \textbf{3.2 points} over strong retrieval baselines, while using substantially smaller tables in the \textbf{50\% configuration}. Overall, by decoupling capacity from compute through efficient memory management, \aname offers a scalable and practical paradigm for future memory-augmented architectures.
\end{abstract}

\begin{document}

\maketitle

\section{Introduction}


Increasing model capacity in large language models usually comes with higher inference-time compute, making further scaling increasingly costly in practice. Token-indexed memory offers an alternative scaling path: instead of relying solely on dense computation, the model can retrieve token-conditioned features from a lookup table, thereby expanding \emph{static} capacity without proportionally increasing compute \citep{mole_ref,scone_ref}. In practice, however, prior lookup-based designs often require very large tables and substantial memory traffic, making these gains difficult to realize efficiently \citep{mole_ref,scone_ref}.

\begin{figure}[t]
\centering
\includegraphics[width=0.75\columnwidth]{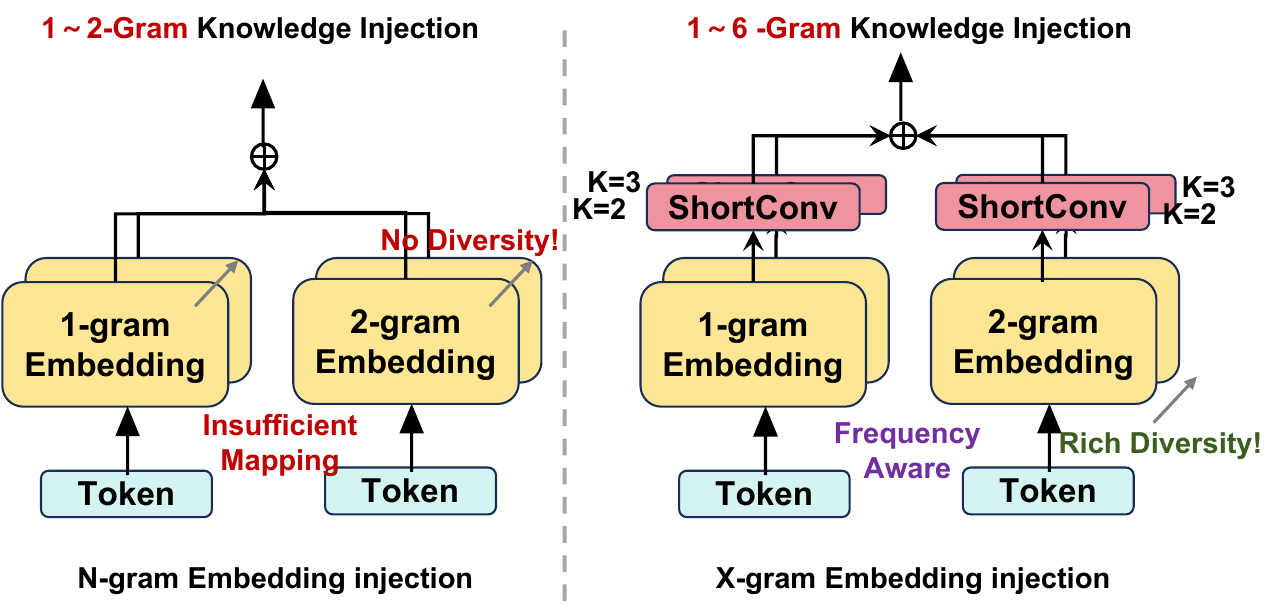}
\caption{Traditional n-gram approaches ignore token frequency, resulting in inefficient parameter allocation—most long-tail slots are under-trained, and slot representations collapse to homogeneous features. In contrast, \aname introduces frequency-aware hybrid routing and diverse contextual feature extraction, enabling efficient parameter usage and richer context modeling.}
\label{fig:xgram}
\vspace{-2mm}
\end{figure}

Yet, in practice, lookup-based scaling shifts the bottleneck from FLOPs to memory traffic, and naive parameter growth can be wasteful.
Prior work often requires tables that are several times larger than the backbone (e.g., MoLE reports $2.4$--$7.4\times$ larger re-parameterized tables), while $n$-gram extensions can demand terabytes of external storage (e.g., $7.6$~TB for $10^9$ $n$-gram embeddings) \citep{mole_ref, scone_ref}.
Such tables rarely fit in HBM; they must be placed in host memory or even SSD, and the resulting parameter traffic is difficult to fully hide behind computation, especially during decoding.
More importantly, empirical gains from enlarging lookup tables often diminish rapidly, suggesting low activation efficiency and information density per byte.

To make this scaling axis practical, we ask a more basic question: what does an \emph{efficient} lookup table need to optimize for under finite data and bandwidth?
We analyze how token-indexed parameters are trained and how their representations behave, and find two consistent regularities.
First, utilization is strongly frequency-driven under Zipf-like statistics: high-frequency items dominate updates and learned activations, while most long-tail rows remain under-trained as the table grows~\cite{yu2025scalingembeddinglayerslanguage}.
Second, scaling capacity by adding parallel fixed slots is not sufficient: different slots are trained on highly overlapping events, encouraging redundant subspaces and rapid saturation~\citep{mole_ref}.
These findings indicate that the bottleneck is not the \emph{lack} of parameters, but the mismatch between table capacity, update density, and representational diversity.

Motivated by this analysis, we propose \aname, a data-aware framework for fine-grained information extraction and injection.
\textbf{ We name it \aname sinceit converts token-indexed 1-gram retrieval into variable-length local x-gram features via multi-scale ShortConv refinement.}
We factor lookup-based augmentation into three stages---vocabulary-to-memory mapping, information extraction, and injection---and design each stage to increase effective capacity without sacrificing stability or system efficiency.

Specifically, \aname makes three targeted design choices to fully exploit the parameter space of lookup memory while keeping training stable and the system offloading-friendly.
First, to mitigate long-tail under-training, \aname introduces \textbf{probability-balanced hybrid hashing}: it assigns dedicated capacity to high-frequency tokens and routes the long tail into shared hash buckets, mitigating hotspot concentration and improving row-level training balance under a fixed budget.
Second, to prevent fixed-slot collapse and encourage diversity, \aname adds a lightweight \textbf{ShortConv} refinement module that enriches static retrieved vectors with multi-scale local structure,\textbf{extracting differentiated $n$-gram features} even when the underlying tables would otherwise become redundant.
Third, to match heterogeneous injection demand across depth and components, \aname uses a \textbf{mixed injection design} centered on attention content pathways, especially the value stream, together with inter-layer residual injection when beneficial; the value pathway provides contextual matching with a smaller injected width, while depth-aware gating modulates injection strength across layers.
Finally, guided by training dynamics, we stabilize the post-injection residual pathway and design a sparse-aware learning-rate schedule for token-indexed parameters, forming an end-to-end system that is both trainable and scalable.

We evaluate \aname with large-scale pretraining across two model scales.
Under identical backbones and training budgets, \aname consistently lowers validation perplexity, improves downstream accuracy, and substantially reduces lookup-table size.
At 0.73B scale, \aname-50\% reaches 48.5 at 1$\times$, beating Retoken (45.0), MoRT (45.8), and Engram (47.2) with a much smaller table, and it scales to 49.1 at 4$\times$.
At 1.15B scale, \aname-50\% reaches 48.2 at 1$\times$ and 50.8 at 4$\times$, surpassing strong baselines by 2.3 points (Table~\ref{table:main-results}).
Moreover, the ShortConv $n$-gram extractor strengthens scaling behavior: as the lookup table grows, \aname continues to gain, whereas baselines often plateau, indicating that additional memory capacity is converted into useful, distinct features rather than redundant slots.
These results, together with our mechanistic analyses, support \aname as a practical lookup-based scaling axis that is trainable under Zipf-like statistics and diversity-preserving under slot scaling.

\section{General Framework }\label{sec:prelim}\label{sec:prelim:framework}

 \paragraph{Generic Injection Framework}
Given a tokenizer $\tau$, an input sequence is tokenized into token IDs
$\mathbf{x}=(x_1,\ldots,x_T)\in V^T$, with embedding matrix
$\mathbf{E}\in\mathbb{R}^{|V|\times d}$.
A Transformer layer maps hidden states $\mathbf{H}_{\ell-1}\in\mathbb{R}^{T\times d}$
to $\mathbf{H}_\ell$ through self-attention and MLP blocks.
We factor lookup-based injection into three operators:
\emph{(i) token-to-parameter mapping} $\mathcal{R}_\phi$, \emph{(ii) information extraction} $\mathcal{T}_\psi$, and \emph{(iii) injection} $\mathcal{I}_s$.
Given token IDs $\mathbf{x}$, each view $m$ retrieves a sequence
\begingroup

\begin{equation}
\mathbf{E}_{\ell}^{(m)}=\mathcal{R}_\phi^{(m)}(\mathbf{x};\mathbf{B}_{\ell}^{(m)}),
\label{eq:ple_map}
\end{equation}
\endgroup
where $\mathbf{B}_{\ell}^{(m)}\in\mathbb{R}^{S\times d_s}$ is a (possibly compressed) lookup table and $\phi$ parameterizes the routing from tokens (or $n$-grams) to physical rows.
An extractor produces refined features
\begingroup

\begin{equation}
\tilde{\mathbf{E}}_{\ell}^{(m)}=\mathcal{T}_\psi^{(m)}(\mathbf{E}_{\ell}^{(m)}),
\label{eq:ple_extract}
\end{equation}
\endgroup
and we fuse views to form a per-layer injection
\begingroup

\begin{equation}
\Delta_{\ell}=\sum_{m=1}^{M_\ell} g_{\ell}^{(m)}\,\tilde{\mathbf{E}}_{\ell}^{(m)},
\label{eq:ple_raw}
\end{equation}
\endgroup
Here, a view denotes an independent retrieval-and-extraction branch with dedicated parameters, while a site denotes the target pathway where the fused signal is injected. The fused signal is injected into a chosen site $s$ by $\mathcal{I}_s$.
We discuss concrete injection sites and contextual matching in Section~\ref{sec:method:inject}.

\paragraph{Budgets.}
Lookup-based augmentation is fundamentally an efficiency problem: the injection design determines how much memory, activation traffic, and compute are spent per token.
We therefore track three budgets:
(i) \textbf{memory budget} (cache size) $\mathcal{C}$, dominated by the lookup table size $S\cdot d_s$ where $S$ is the number of physical rows and $d_s$ is the injected dimensionality at injection site $s$;
(ii) \textbf{activation budget} $\mathcal{A}$, the per-token activation footprint of the injected pathway, proportional to $d_s$;
Here, activation refers to the memory footprint of intermediate tensors saved for backpropagation, rather than nonlinear activation functions.
and (iii) \textbf{compute budget} $\mathcal{F}$, the extra per-token FLOPs induced by lookup aggregation, feature extraction, which also scales with $d_s$.
We refer to improvement per unit $\mathcal{C}$ as \textbf{parameter efficiency}, and improvement per unit $\mathcal{A}$ as \textbf{activation efficiency}.

\paragraph{Objective.}
Given a backbone $f_\theta$ and an injection channel parameterized by $\varphi$, we train the augmented model $f_{\theta,\varphi}$ to minimize the expected language-modeling loss under explicit budgets:
\begingroup

\begin{equation}
\begin{split}
\min_{\theta,\varphi}\ 
\mathbb{E}_{(\mathbf{x},y)\sim\mathcal{D}}
\left[\ell\!\left(f_{\theta,\varphi}(\mathbf{x}),y\right)\right] ~~~~~~~~~~~~~~~~~~~~~~~~~~ \\
\quad \text{s.t.}\quad 
\mathcal{C}(\varphi)\le \mathcal{C}_0,
\mathcal{A}(\varphi)\le \mathcal{A}_0,\ 
\mathcal{F}(\varphi)\le \mathcal{F}_0.
\end{split}
\label{eq:ple_obj}
\end{equation}
\endgroup

\paragraph{How \aname instantiates Eq.~(\ref{eq:ple_obj}).}
Our method is designed to improve quality while staying within the three explicit budgets.
For the \textbf{memory budget} $\mathcal{C}$, we reduce the effective table size $S$ via probability-balanced VIP+hash routing with alias mixing, so more updates concentrate on trainable rows (Section~\ref{sec:method:routing} and Table~\ref{tab:abla_hash}).
For the \textbf{activation budget} $\mathcal{A}$, we inject into attention content pathways with a small injected dimensionality (typically $d_s{=}d_{\mathrm{kv}}$) and use fewer views than prior multi-slot designs (Section~\ref{sec:method:inject} and Table~\ref{tab:budget_alignment}).
For the \textbf{compute budget} $\mathcal{F}$, we use a lightweight depthwise ShortConv with a small set of kernels and avoid expensive score-path injection by default (Section~\ref{sec:method:shortconv} and Table~\ref{tab:inj_budgets}).
These choices are reflected in our budget-aligned experimental protocol (Section~\ref{sec:exp_setup}) and the analytic budget summaries in Tables~\ref{tab:inj_budgets} and \ref{tab:budget_alignment}.

\paragraph{Other methods as framework instantiations.}
The same three-operator view also clarifies how prior token-indexed methods differ.
Table~\ref{tab:framework_instantiations} writes Retoken, MoRT, Engram, and \aname by directly substituting their routing, extraction, and injection choices into the framework.
This comparison highlights the key distinction: fixed lookup and explicit $n$-gram tables scale memory by materializing more rigid slots, whereas \aname keeps routing compact and uses learnable convolutional extraction to adaptively compose local multi-token information.

\begin{table*}[t]
\centering
\caption{
Representative methods written as direct instantiations of the three-operator framework.
Dominant per-layer memory and incremental per-token FLOPs omit scalar gates and layer-wise constants. $\mathrm{SC}_{k}$ means SwiGLUShortConv with kernel size $k$.
}
\label{tab:framework_instantiations}
\vspace{0.5mm}
\setlength{\tabcolsep}{3pt}
\footnotesize
\resizebox{\textwidth}{!}{%
\begin{tabular}{@{}l p{0.60\textwidth} c c@{}}
\toprule
\textbf{Method} & \textbf{Framework form} & \textbf{Dominant memory} & \textbf{Extra FLOPs/token} \\
\midrule
\textbf{Retoken} &
$\begin{aligned}[t]
\Delta^{\mathrm{ffn}}_{\ell,t}\leftarrow
\Delta^{\mathrm{ffn}}_{\ell,t}\odot
\prod_{m=1}^{M}\Bigl(
1+\mathbf{a}_{\ell}^{(m)}\odot
\mathrm{L2Norm}(\mathbf{B}_{\ell}^{(m)}[x_t])
\Bigr)
\end{aligned}$ &
$M|\mathcal{V}|d+Md$ &
$O(Md)$ \\
\addlinespace[2mm]
\textbf{MoRT} &
$\begin{alignedat}[t]{2}
\mathbf{h}_t \leftarrow \mathbf{h}_t
&\;+\; \frac{1}{\sqrt{2N_\ell}}\,
\mathbf{s}_{\ell}\odot \mathrm{L2Norm}\!\Biggl(
\sum_{m\in \mathrm{TopK}(\mathbf{z}_t)}
\alpha_{t,m}\mathbf{B}_{\ell}^{(m)}[x_t]
\Biggr),\\
\alpha_{t,m} &\,=\, 
\frac{\sigma((\mathbf{z}_t)_m)}
{\sum_{r\in \mathrm{TopK}(\mathbf{z}_t)} \sigma((\mathbf{z}_t)_r)}, ~~~~~~\mathbf{z}_t =\, \mathbf{W}_r \mathbf{h}_t, 
\end{alignedat}$ &
$M|\mathcal{V}|d+dM+d$ &
$O(Md)$ \\
\addlinespace[2mm]
\textbf{Engram} &
$\begin{aligned}[t]
\mathbf{h}_t \leftarrow \mathbf{h}_t
&+ g_t \mathbf{W}_v
\bigl[\mathbf{B}_{\ell}^{(n,k)}[h_{n,k}(x_{t-n+1:t})]\bigr]_{n\in\mathcal{N},\,k\in[K]} \\
&+ \mathrm{ShortConv}\!\Bigl(
g_t \mathbf{W}_v
\bigl[\mathbf{B}_{\ell}^{(n,k)}[h_{n,k}(x_{t-n+1:t})]\bigr]_{n\in\mathcal{N},\,k\in[K]}
\Bigr)
\end{aligned}$ &
$\sum_{n\in\mathcal{N}}\sum_{k=1}^{K} S_{n,k} d_{n,k} + 2d_ed$ &
$O(2d_ed + (k+1)d)
$ \\
\addlinespace[2mm]
\textbf{\aname} &
$\begin{aligned}[t]
\mathbf{V}_{\ell,t} &\leftarrow \mathbf{V}_{\ell,t}+\Delta_{\ell,t}
\quad \text{and} \quad
\mathbf{h}_{\ell-1,t} \leftarrow \mathbf{h}_{\ell-1,t}+\Delta_{\ell,t}, \\
\Delta_{\ell,t}
&= \frac{1}{\sqrt{M_\ell}} \sum_m g_{\ell}^{(m)}(u)
\left(
\mathbf{e}_{\ell,t}^{(m)}
+
\mathrm{SC}_{k^{(m)}}\!\bigl(\mathrm{RMSNorm}(\mathbf{e}_{\ell}^{(m)})\bigr)_t
\right), \mathbf{e}_{\ell,t}^{(m)}
= \mathcal{R}_{\mathrm{hash}}^{(m)}(x_t)
\end{aligned}$ &
$\rho|\mathcal{V}|\sum_s M_s d_s$ &
$O(M(K+2k+1)d_s)$ \\
\bottomrule
\end{tabular}%
}
\vspace{-2mm}
\end{table*}


\section{Observation}\label{sec:observation}

\label{sec:observation_main}
We next present observations that explain why naive lookup scaling saturates, and which guide the design of \aname.
To ground the following analysis, we use a simple per-layer embedding (PLE) baseline, which aggregates token-indexed retrieved features and injects them at each layer.
In the PLE model, each layer aggregates $M_\ell$ token-indexed views and injects their normalized sum into the layer:
\begingroup

\begin{equation}
\Delta_{\ell}=\frac{1}{\sqrt{M_\ell}}\sum_{m=1}^{M_\ell} g_{\ell}^{(m)}\,\tilde{\mathbf{E}}_{\ell}^{(m)}.
\label{eq:obs_inj}
\end{equation}
\endgroup
Unless stated otherwise, we report statistics after training for 41.4B tokens; metric definitions are given in Appendix~\ref{appobs:metrics}.

\begin{figure}[tbp]
\centering
\IfFileExists{figures/obs.pdf}{%
  \includegraphics[width=\columnwidth]{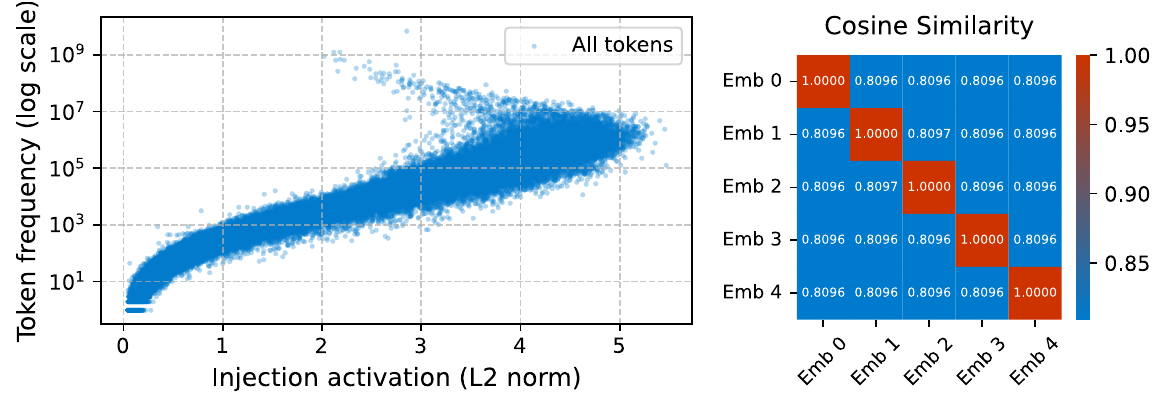}%
  \begin{subfigure}[b]{0pt}
    \phantomcaption\label{fig:obs_trainability}
  \end{subfigure}%
  \begin{subfigure}[b]{0pt}
    \phantomcaption\label{fig:obs_cosine}
  \end{subfigure}
}{%
  \fbox{\parbox{\columnwidth}{\vspace{10mm}\centering Placeholder: figures/obs.pdf\vspace{10mm}}}%
}
\caption{Left: Long-tail trainability of token-indexed parameters. Right: Cosine similarity among parallel Embeddings.}
\label{fig:obs_main}
\vspace{-3mm}
\end{figure}

\paragraph{Utilization follows token frequency}\label{obs:trainability}
We investigate whether larger token-indexed tables provide genuinely usable capacity under realistic data budgets, noting that due to the long-tail distribution of token and $n$-gram frequencies, most table rows receive very few updates and remain near initialization. To measure effective utilization, we use each token’s average activation magnitude across all occurrences and layers.
Figure~\ref{fig:obs_trainability} reveals two key effects: (i) activation magnitude increases strongly with frequency, indicating that learned lookup features are concentrated on the head; and (ii) the fraction of sufficiently-updated rows drops sharply as the table grows, so uniform scaling primarily adds cold, under-trained parameters.
This observation turns table scaling into an allocation problem: we need to concentrate capacity where updates are dense, while compressing the tail in a way that preserves update efficiency. The row-level statistics in Appendix~\ref{app:hash_diagnostics} show that frequency-aware hashing reduces row-hit concentration and markedly flattens within-table training inequality.

\paragraph{Fixed-slot scaling saturates via redundancy}\label{obs:redundancy}
We then ask why simply adding more parallel lookup slots often yields diminishing returns even when memory budget increases.
In fixed-slot designs, each slot sees almost the same token occurrences and thus receives highly correlated gradients; as a result, ``more slots'' does not reliably create ``more directions'' in representation space.
We diagnose this by measuring pairwise cosine similarity between retrieved embeddings across slots (Appendix~\ref{appobs:metrics}).
Figure~\ref{fig:obs_cosine} shows that slot representations are highly similar, indicating subspace collapse and limited angular diversity.
The implication is that parameter count alone is not a sufficient scaling axis: without a mechanism that \emph{breaks symmetry} across slots, additional tables mostly replicate existing features.

These findings indicate that lookup-based scaling is limited by frequency-agnostic allocation and slot redundancy, which our method overcomes via frequency-aware allocation and diversity-enhancing feature extraction.

\section{Methodology}
\label{sec:method}
We introduce \aname, a token-indexed injection framework that enables Transformers to scale more efficiently by allocating parameters in a fine-grained, frequency-aware manner and by enhancing the embedding scaling axis through local feature extraction for diverse, effective memory augmentation.

\subsection{Frequency-Aware Hash Mapping}
\label{sec:method:routing}

Lookup tables are typically trained under access patterns that approximately follow a Zipfian distribution. If capacity is allocated uniformly across the entire vocabulary, many physical rows will remain associated with tokens that are rarely accessed and rarely updated, leading to low parameter utilization. To address this issue, we introduce a frequency-aware mapping mechanism: under a fixed physical budget, more capacity is allocated to the high-frequency region, while the remaining tokens are compressed in a controlled manner, improving overall update efficiency and trainability.

Concretely, we first estimate the empirical frequency of each token from a sampled corpus and denote it by $p(\omega)$. To mitigate the extreme imbalance induced by the original Zipfian distribution, we define a smoothed mass
\begingroup

\begin{equation*}
s(\omega)=p(\omega)^{\alpha}, \qquad 0<\alpha\le 1,
\end{equation*}
\endgroup
and construct a frequency-aware bucketing scheme accordingly. We first extract a small set of the most frequent tokens as the head set $\mathcal{V}_{\mathrm{vip}}$, and assign dedicated physical rows to them in a reserved VIP region. The motivation is to avoid the severe imbalance that arises when applying a uniform bucketing procedure directly to the raw Zipfian distribution: a few extremely frequent tokens may occupy several nearly empty head buckets, while a large number of tail tokens are compressed into only a few highly crowded buckets at the end. After removing these VIP tokens, the remaining frequency distribution becomes more balanced, which is more suitable for subsequent balanced bucketing and capacity allocation.

For the remaining tokens outside $\mathcal{V}_{\mathrm{vip}}$, we partition them into $B$ logical buckets $\{\mathcal{G}_b\}_{b=1}^{B}$ according to the smoothed mass, such that the total mass of each bucket is approximately balanced:
\begingroup

\begin{equation}
\sum_{\omega\in \mathcal{G}_b} p(\omega)^{\alpha}
\approx
\frac{1}{B}\sum_{\omega\in\mathcal{V}\setminus \mathcal{V}_{\mathrm{vip}}} p(\omega)^{\alpha}.
\label{eq:mass_balance}
\end{equation}
\endgroup
Let $S$ denote the compressed physical table size. The compression ratio is then defined as $\rho = S/|\mathcal{V}|$.

\begin{figure}[tbp]
\centering
\includegraphics[width=0.7\columnwidth]{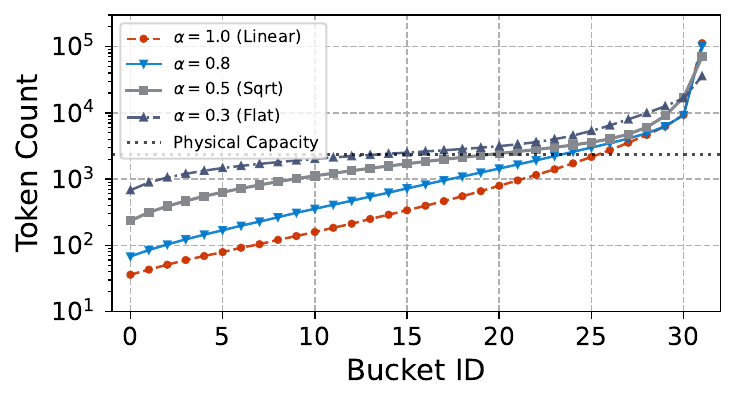}
\caption{Frequency-aware hash mapping combines VIP reservation, balanced logical bucketing, and bucket-local mapping to allocate capacity more effectively under a fixed physical budget.}
\label{fig:token_distribution}
\vspace{-4mm}
\end{figure}

After the logical grouping is completed, we further map tokens to concrete physical rows. For tokens in $\mathcal{V}_{\mathrm{vip}}$, we assign dedicated rows in the VIP region through deterministic offsets; when needed, a small number of additional rows can also be assigned and combined with decayed weights. For the remaining tokens, we first determine the available physical subregion according to the assigned logical bucket, and then perform intra-bucket mapping. If a bucket is sparse, tokens are mapped directly to physical rows according to their intra-bucket order, and the unused rows can be recycled as alias rows. If a bucket is dense and collisions are unavoidable, we adopt multi-path local hashing within the bucket, generating a small number of retrieval paths for the same token and aggregating the corresponding results to reduce collision noise. In this way, different frequency regions can use different collision-handling strategies, leading to a more balanced distribution of collision and update pressure under a fixed budget.

Accordingly, the retrieved representation of token $\omega$ at layer $\ell$ and branch $m$ can be written in a compact form as
\begingroup

\begin{equation}
\mathbf{e}_{\ell}^{(m)}(\omega)
=
\sum_{j\in \mathcal{A}(\omega)}
c_{\omega,j}\,
\sigma\!\bigl(a_{\ell,j}^{(m)}\bigr)\,
\mathbf{B}_{\ell}^{(m)}[j],
\label{eq:alias_retrieval}
\end{equation}
\endgroup
where $\mathcal{A}(\omega)$ denotes the access list of a small number of physical rows visited by token $\omega$, $c_{\omega,j}$ denotes the effective aggregation coefficient associated with physical row $j$, $a_{\ell,j}^{(m)}$ is a row-wise learnable parameter associated with physical row $j$, and $\sigma(\cdot)$ denotes the Sigmoid activation function. For VIP tokens and ordinary tokens in sparse buckets, $\mathcal{A}(\omega)$ consists of a deterministically assigned base row together with optional additional rows; for ordinary tokens in dense buckets, $\mathcal{A}(\omega)$ consists of the physical rows visited by multi-path local hashing. Overall, the proposed mechanism combines head reservation with adaptive intra-bucket mapping, making the distribution of collision and update pressure more balanced across tokens of different frequencies under a fixed budget, and thereby yielding a lookup table representation that is easier to train. Additional details are provided in the appendix.

\subsection{\aname Extraction with Gated ShortConv} 
\label{sec:method:shortconv}
\begin{figure*}[!tp]
\centering
\includegraphics[width=\textwidth]{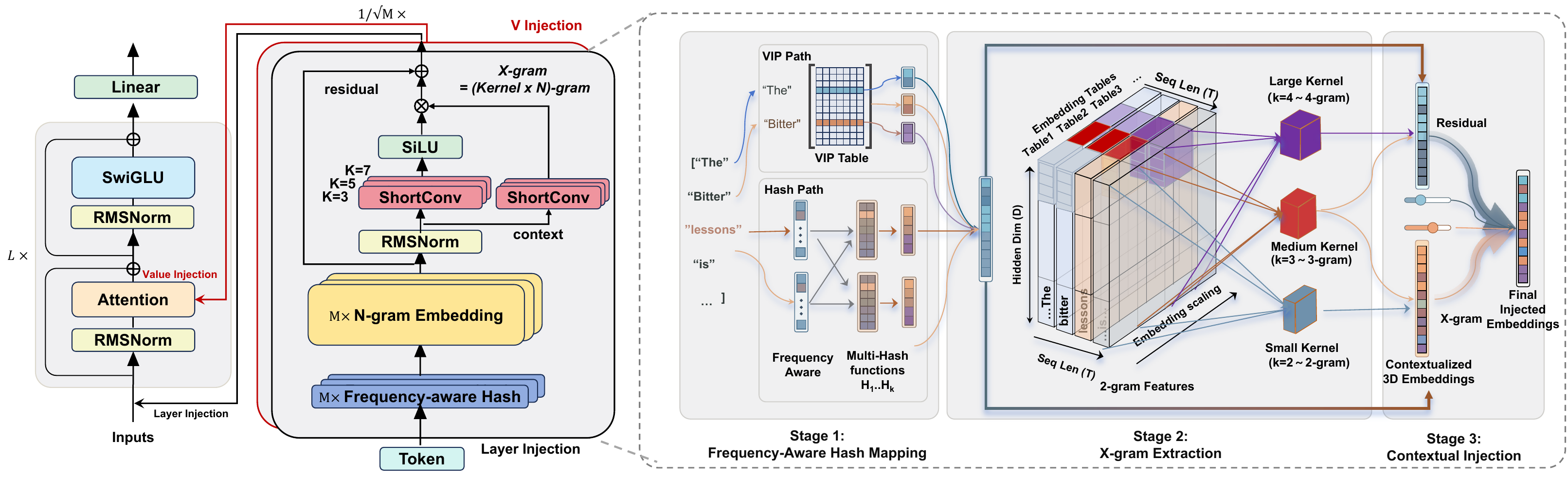}
\caption{
Overview of \aname.
We retrieve token-indexed vectors from a compact lookup memory via frequency-aware routing (Stage 1), extract local $n$-gram features through a gated ShortConv module (normalized for stability; Section~\ref{sec:method:opt}) (Stage 2), and inject the resulting signals into the attention value pathway and/or inter-layer residuals with depth-aware gating (Stage 3).
}\label{fig:main}
\end{figure*}

Lookup retrieval is context-independent, but whether a retrieved vector is useful depends on local context.
Moreover, without a view-specific nonlinear extractor, parallel memories can collapse to redundant subspaces (Appendix~\ref{appobs:linear}).
We therefore refine each retrieved sequence with a lightweight, normalized, gated \emph{multi-scale} ShortConv that extracts local $x$-gram cues and promotes diversity across views.

Let $\mathbf{E}_{\ell}^{(m)}\in\mathbb{R}^{T\times d_s}$ denote the retrieved sequence at layer $\ell$ for view $m$.
We first normalize inputs to stabilize the multiplicative gate under sparse updates:
\begingroup

\begin{equation}
\bar{\mathbf{E}}_{\ell}^{(m)}=\mathrm{RMSNorm}(\mathbf{E}_{\ell}^{(m)}).
\label{eq:shortconv_norm}
\end{equation}
\endgroup

Each view uses a kernel size $k^{(m)}$, and different views collectively provide multi-scale coverage. For each view $m$, we apply causal padding and depthwise (channel-wise) convolutions for a content branch and a gate branch:

\begingroup

\begin{equation}
\begin{aligned}
\mathbf{C}_{\ell}^{(m)}
&=
\mathrm{DWConv1D}_{\psi_c^{(m)},\,k^{(m)}}
\!\left(
\bar{\mathbf{E}}_{\ell}^{(m)}
\right)\odot\;
\mathrm{SiLU}\!\left(
\mathrm{DWConv1D}_{\psi_g^{(m)},\,k^{(m)}}
\!\left(
\bar{\mathbf{E}}_{\ell}^{(m)}
\right)
+\mathbf{b}^{(m)}
\right),
\end{aligned}
\label{eq:shortconv_multiscale}
\end{equation}
\endgroup

We fuse the extracted feature with a residual to preserve a stable anchor:
\begingroup

\begin{equation}
\tilde{\mathbf{E}}_{\ell}^{(m)}=\mathbf{E}_{\ell}^{(m)}+\mathbf{C}_{\ell}^{(m)}.
\label{eq:shortconv_residual_scale}
\end{equation}
\endgroup

Different views use different kernel sizes, so the resulting extractors collectively provide multi-scale coverage and expand the receptive field from token identity to local $x$-grams.
The gate acts as a context-selective filter that suppresses collision-induced noise.
View-specific kernels break symmetry across parallel memories, increasing diversity before fusion.
Additional rationale is provided in Appendix~\ref{app:shortconv_rationale}.

\subsection{Contextual Injection}
\label{sec:method:inject}

\paragraph{Injection sites under budget constraints.}
Injection location determines how much of the lookup signal is exposed to contextual matching, and how much budget is spent per token.
Under grouped-query attention (GQA) with $h$ query heads and $h_{\text{kv}}$ key/value heads, the key/value width is $d_{\text{kv}}\triangleq (h_{\text{kv}}/h)\,d$ and typically $d_{\text{kv}}\ll d$.
This makes injecting into the attention value stream particularly attractive: it directly augments the content pathway while keeping the activation footprint small (Table~\ref{tab:inj_budgets}).   

\begin{algorithm}[t]
  \small
  \caption{\aname: Extraction and Injection}
  \label{alg:method}
  
  \begin{algorithmic}[1]
  \renewcommand{\algorithmiccomment}[1]{\hfill \textit{// #1}}
  
  \REQUIRE $\mathbf{x}$, step $u$, layer $\ell$, views $M_\ell$, tables $\mathbf{B}_{\ell}^{(m)}$
  \REQUIRE weights $\pi, a, \psi$
  \FOR{$m=1$ to $M_\ell$ }
      \FOR{$t=1$ to $T$}
          \STATE $\mathbf{e}_t \leftarrow  \mathrm{Lookup}_\ell^{(m)}(x_t)$  \COMMENT{Frequency-aware Retrieval}
      \ENDFOR
      \STATE $\mathbf{E}_{\ell}^{(m)} \leftarrow [\mathbf{e}_1;\ldots;\mathbf{e}_T]$
      \STATE $\hat{\mathbf{E}} \leftarrow \mathrm{RMSNorm}(\mathbf{E}_{\ell}^{(m)})$ \COMMENT{ShortConv Extraction}
      \STATE $\tilde{\mathbf{E}}_{\ell}^{(m)} \leftarrow \mathbf{E}_{\ell}^{(m)} + \mathrm{ShortConv}(\hat{\mathbf{E}})$
  \ENDFOR
  \STATE $\Delta_{\ell} \leftarrow \frac{1}{\sqrt{M_\ell}} \sum_{m} g_{\ell}^{(m)}(u)\,\tilde{\mathbf{E}}_{\ell}^{(m)}$ \COMMENT{Aggregation \& Injection}
  \STATE $\mathbf{Q}, \mathbf{K}, \mathbf{V} \leftarrow \mathrm{Proj}(\mathbf{H}_{\ell-1})$ \COMMENT{Standard Attention}
  \STATE $\mathbf{V}_\ell \leftarrow \mathbf{V} + \Delta_\ell$ \COMMENT{Value injection}
  \STATE $\mathbf{O}_\ell \leftarrow \mathrm{Attention}(\mathbf{Q}, \mathbf{K}, \mathbf{V}_\ell)$
  \RETURN $\mathbf{O}_\ell$ (or $\mathbf{H}_\ell \leftarrow \mathrm{Block}(\mathbf{H}_{\ell-1} + \Delta_\ell)$)
\end{algorithmic}
\end{algorithm}

\paragraph{Attention-value injection.}
We inject the fused lookup signal $\Delta_\ell$ into the attention value stream as
\begingroup

\begin{equation}
\mathbf{V}_\ell \leftarrow \mathbf{H}_{\ell-1}\mathbf{W}_\ell^{V} + \Delta_\ell,
\label{eq:v_inject}
\end{equation}
\endgroup
so attention weights provide contextual matching over the injected semantics without perturbing similarity scores.

\paragraph{Inter-layer residual injection.}
When the long tail is aggressively compressed, we optionally add an inter-layer residual path,
\begingroup

\begin{equation}
\mathbf{H}_\ell \leftarrow \mathrm{Block}_\ell(\mathbf{H}_{\ell-1} + \Delta_\ell),
\label{eq:interlayer_inject}
\end{equation}
\endgroup
which acts as a stable identity anchor that is insensitive to the internal attention/MLP parameterization.

\subsection{Training Stability and Optimization}
\label{sec:method:opt}

Token-indexed parameters are updated sparsely and can destabilize training if their scale and learning dynamics are mismatched with the backbone.
We control stability with three scale rules and a sparse-aware optimization scheme; detailed derivations are provided in Appendix~\ref{app:stability}.

We normalize multi-view fusion to keep the injected residual approximately invariant to the number of views:
\begingroup

\begin{equation}
\Delta_{\ell}=\frac{1}{\sqrt{M_\ell}}\sum_{m=1}^{M_\ell} g_{\ell}^{(m)}(u)\,\tilde{\mathbf{E}}_{\ell}^{(m)}.
\label{eq:triop}
\end{equation}
\endgroup
We use a depth-aware warm-gated schedule to avoid abrupt amplification at depth and during early training:
\begingroup

\begin{equation}
g_{\ell}^{(m)}(u)=\lambda_{\ell}^{(m)}\cdot \sqrt{\ell+1}\cdot w_{\text{warmup}}(u),
\label{eq:gate_rule}
\end{equation}
\endgroup
where $u$ is the optimization step, $\lambda_{\ell}^{(m)}$ is learnable, and $w_{\text{warmup}}(u)$ ramps from 0 to 1 during early training.
Finally, we stabilize multiplicative gating by normalizing the retrieved sequence before ShortConv (Eq.~(\ref{eq:shortconv_norm})--Eq.~(\ref{eq:shortconv_multiscale})).

For optimization, we treat lookup-table parameters as a sparse group and apply a larger effective step size than the backbone to compensate for low update frequency.
Concretely, we scale the lookup learning rate as a function of expected row hit rate, which depends on table size $S$ (equivalently $\rho|\mathcal{V}|$); a practical rule is $\eta_{\text{lookup}}\propto \sqrt{S}$ up to clipping, with further refinements in Appendix~\ref{app:stability}.

\subsection{System Efficiency and Offloading}
\begin{table}[t]
\centering
\caption{Budget accounting of \aname injection sites. $S=|\mathcal{V}|$, $\kappa = (K+2k+1)$.}
\label{tab:inj_budgets}
\small 
\setlength{\tabcolsep}{6pt} 
\begin{tabular}{lccc}
\toprule
\textbf{Injection Site} & $\Delta\text{Params}$ & $\Delta\text{Dims}$ & $\Delta\text{FLOPs}$ \\
\midrule
Q (pathway) & $S d_{\text{kv}}$ & $d_{\text{kv}}$ & $\kappa d_{\text{kv}}$ \\
K (pathway) & $S d_{\text{kv}}$ & $d_{\text{kv}}$ & $\kappa d_{\text{kv}}$ \\\
V (content) & $S d_{\text{kv}}$ & $d_{\text{kv}}$ & $\kappa d_{\text{kv}}$ \\
O (output) & $S d$ & $d$ & $\kappa d$ \\
Inter-layer & $S d$ & $d$ & $\kappa d$ \\
V + hash & $\rho S d_{\text{kv}}$ & $d_{\text{kv}}$ & $\kappa d_{\text{kv}}$ \\
Inter-layer + hash & $\rho S d$ & $d$ & $\kappa d$ \\
\bottomrule
\addlinespace
\end{tabular}
\end{table}
Token-indexed memory can be large, so we quantify incremental overhead to keep the scaling axis practical.
Per view, memory scales as $\Delta\mathcal{C}=S d_s$ (with $S=|\mathcal{V}|$ or $S=\rho|\mathcal{V}|$), per-token activation is $\Delta\mathcal{A}=d_s$, and compute is $\Delta\mathcal{F}\approx (K+2k+1)d_s$ (Table~\ref{tab:inj_budgets}).
Thus, smaller injected width (e.g., value-stream injection under GQA) reduces activation/compute and, under fixed cache $\mathcal{C}_0=\rho|\mathcal{V}|d_s$, permits a larger $\rho$ (fewer collisions).

Offloading is feasible because retrieval depends only on token IDs and static maps.
Let $U$ be the number of unique token IDs in a batch and $U_{\ell}\le \min\{S,KU\}$ the unique physical rows referenced at injected layer $\ell$; the prefetch cost is
\begingroup

\begin{equation}
T_{\text{comm}}\ \approx\ \sum_{\ell\in\mathcal{L}_{\text{inj}}}\ \frac{U_{\ell}\,d_s\,b}{\mathrm{BW}_{\text{host}\to\text{dev}}},
\label{eq:offload_comm}
\end{equation}
\endgroup
so compression and selective injection reduce both $\mathcal{C}$ and $U_{\ell}$, improving offloading feasibility in principle.

\FloatBarrier

\section{Experiments}

\begin{table*}[!t]
\centering
\caption{Main downstream results at two backbone scales. Each block compares methods under matched training budgets and comparable token-indexed parameter sizes; for \aname, the 1$\times$/2$\times$/4$\times$ multipliers denote fixed $h/v$ view configurations summarized in Appendix Table~\ref{tab:main_table_configs}.}
\vskip 0.05in
\renewcommand{\arraystretch}{0.85}
\resizebox{\textwidth}{!}{%
\setlength{\tabcolsep}{3.0pt}
\scriptsize
\begin{tabular}{llcccccccccccc}
\toprule
\textbf{Method} & \textbf{Cfg} & \textbf{\# Params} & \textbf{SciQ} & \textbf{PIQA} & \textbf{WG} & \textbf{ARC-E} & \textbf{ARC-C} & \textbf{Hella.} & \textbf{S.QA} & \textbf{BoolQ} & \textbf{OBQA} & \textbf{MMLU} & \textbf{Avg.} \\
\midrule
\multicolumn{14}{l}{\textbf{0.73B Scale}} \\
\addlinespace[1pt]
Baseline &  & 0.73B & 78.4 & 60.9 & 51.3 & 41.4 & 25.3 & 35.8 & 39.8 & 60.8 & 27.6 & 25.4 & 44.7 \\
\midrule
Retoken & \multirow{5}{*}{1$\times$} & 0.73B+2.33B & 77.5 & 62.4 & 48.1 & 46.1 & 25.5 & 36.3 & 41.0 & \textbf{61.2} & 28.0 & 24.0 & 45.0 \\
MoRT & & 0.73B+2.33B & 74.3 & 63.3 & \textbf{54.1} & 47.9 & 25.8 & 35.5 & 42.2 & 60.5 & 28.0 & 26.3 & 45.8 \\
Engram & & 0.73B+2.33B & 77.9 & 65.6 & 51.6 & 53.5 & 27.8 & 40.2 & 41.7 & 60.0 & 27.8 & 25.7 & 47.2 \\
\aname-50\% & & 0.73B+1.17B & \textbf{82.9} & \textbf{67.5} & 52.1 & \textbf{56.7} & 27.5 & \textbf{41.5} & 42.6 & 60.4 & \textbf{28.8} & 24.9 & \textbf{48.5} \\
\aname-100\% & & 0.73B+2.33B & 81.0 & 65.9 & 49.6 & 54.0 & \textbf{27.9} & 40.6 & \textbf{42.8} & 60.1 & 26.8 & \textbf{26.4} & 47.5 \\
\midrule
Retoken & \multirow{5}{*}{2$\times$} & 0.73B+4.67B & 78.8 & 65.7 & 49.9 & 51.2 & 26.2 & 36.8 & 42.2 & 60.4 & 29.0 & 25.9 & 46.6 \\
MoRT & & 0.73B+4.67B & 78.2 & 62.5 & 52.6 & 47.9 & 25.0 & 36.0 & 41.9 & \textbf{61.6} & 28.8 & 24.7 & 45.9 \\
Engram & & 0.73B+4.67B & 78.1 & 67.5 & 50.8 & 53.5 & 28.1 & 41.8 & \textbf{43.4} & 59.5 & 29.6 & 24.6 & 47.7 \\
X-GRAM-50\% & & 0.73B+2.33B & \textbf{83.5} & 68.4 & \textbf{54.7} & 53.0 & \textbf{29.7} & 42.2 & 41.5 & 59.1 & 27.4 & \textbf{27.3} & 48.7 \\
X-GRAM-100\% & & 0.73B+4.67B & \textbf{83.5} & \textbf{68.9} & 53.7 & \textbf{59.1} & 28.0 & \textbf{43.2} & 42.8 & 60.8 & \textbf{30.4} & 26.4 & \textbf{49.7} \\
\midrule
Retoken & \multirow{5}{*}{4$\times$} & 0.73B+9.33B & 80.4 & 63.4 & 50.3 & 48.8 & 25.3 & 37.2 & 41.8 & 58.0 & 25.8 & 25.4 & 45.6 \\
MoRT & & 0.73B+9.33B & 75.0 & 63.1 & 52.2 & 49.6 & 25.4 & 36.5 & 42.2 & 59.9 & 29.6 & 25.5 & 45.9 \\
Engram & & 0.73B+9.33B & 73.9 & 62.6 & 51.4 & 43.5 & 27.6 & 39.2 & 40.6 & \textbf{60.9} & 29.0 & \textbf{27.3} & 45.6 \\
X-GRAM-50\% & & 0.73B+4.67B & 82.8 & \textbf{70.2} & \textbf{52.8} & 58.1 & \textbf{29.9} & 41.8 & 42.5 & 57.6 & \textbf{32.0} & 23.7 & 49.1 \\
X-GRAM-100\% & & 0.73B+9.33B & \textbf{84.4} & 69.6 & 51.5 & \textbf{58.4} & 28.1 & \textbf{44.0} & \textbf{43.5} & 59.8 & 30.2 & 25.8 & \textbf{49.5} \\
\midrule
\multicolumn{14}{l}{\textbf{1.15B Scale}} \\
\addlinespace[1pt]
Baseline &  & 1.15B & 82.0 & 66.2 & 52.2 & 53.2 & 28.7 & 41.0 & 42.8 & 54.4 & 29.2 & 23.9 & 47.4 \\
\midrule
Retoken & \multirow{5}{*}{1$\times$} & 1.15B+3.73B & 81.2 & 67.3 & \textbf{52.6} & 52.3 & 26.8 & 41.6 & 42.2 & 56.6 & 28.6 & 26.0 & 47.5 \\
MoRT & & 1.15B+3.73B & 81.0 & 66.0 & 51.7 & 49.3 & 28.6 & 41.2 & 41.6 & \textbf{59.9} & 28.6 & 24.8 & 47.3 \\
Engram & & 1.15B+3.73B & 77.5 & 65.8 & 51.0 & 49.1 & 29.0 & 42.4 & \textbf{43.1} & 58.8 & 28.0 & 24.4 & 46.9 \\
\aname-50\% & & 1.15B+1.87B & \textbf{83.4} & \textbf{68.8} & 50.5 & 56.3 & \textbf{30.6} & \textbf{45.0} & 42.5 & 51.9 & 28.6 & 24.5 & 48.2 \\
\aname-100\% & & 1.15B+3.73B & 82.8 & 68.4 & 51.2 & \textbf{57.7} & 30.5 & \textbf{45.0} & 42.7 & 59.3 & \textbf{30.8} & \textbf{26.4} & \textbf{49.5} \\
\midrule
Retoken & \multirow{5}{*}{2$\times$} & 1.15B+7.47B & 81.5 & 64.4 & 51.3 & 50.7 & 27.0 & 41.1 & \textbf{43.8} & 56.0 & 28.6 & 25.1 & 47.0 \\
MoRT & & 1.15B+7.47B & 80.5 & 66.2 & 50.0 & 55.1 & 27.4 & 40.4 & 42.6 & \textbf{60.0} & 30.4 & \textbf{25.2} & 47.8 \\
Engram & & 1.15B+7.47B & 78.7 & 64.6 & 50.6 & 48.6 & 29.3 & 43.1 & 43.0 & 57.3 & 27.4 & 24.8 & 46.7 \\
X-GRAM-50\% & & 1.15B+3.73B & \textbf{84.8} & \textbf{69.9} & \textbf{53.1} & \textbf{60.5} & 29.9 & 46.5 & \textbf{43.8} & \textbf{60.0} & 31.0 & 25.1 & \textbf{50.5} \\
X-GRAM-100\% & & 1.15B+7.47B & 84.2 & 69.6 & 52.3 & 59.5 & \textbf{30.2} & \textbf{46.6} & 42.5 & 59.4 & \textbf{31.6} & 22.4 & 49.8 \\
\midrule
Retoken & \multirow{5}{*}{4$\times$} & 1.15B+14.94B & 78.6 & 63.5 & 50.0 & 47.9 & 28.8 & 41.1 & 41.6 & 59.0 & 28.4 & 26.0 & 46.5 \\
MoRT & & 1.15B+14.94B & 82.7 & 66.1 & 52.0 & 55.6 & 27.9 & 41.1 & \textbf{44.7} & \textbf{61.8} & 27.2 & 25.1 & 48.4 \\
Engram & & 1.15B+14.94B & 78.0 & 67.6 & 50.8 & 51.6 & 30.7 & 45.6 & 43.1 & 61.4 & 29.8 & \textbf{26.1} & 48.5 \\
X-GRAM-50\% & & 1.15B+7.47B & \textbf{86.0} & 70.2 & \textbf{53.7} & 60.9 & 31.8 & 47.4 & 43.7 & 57.3 & \textbf{33.4} & 23.6 & \textbf{50.8} \\
X-GRAM-100\% & & 1.15B+14.94B & 85.8 & \textbf{70.8} & 52.9 & \textbf{61.1} & \textbf{32.2} & \textbf{47.9} & 43.7 & 57.1 & 30.8 & 25.8 & \textbf{50.8} \\
\bottomrule
\end{tabular}%
 }
\label{table:main-results}
\end{table*}

\subsection{Experimental Setup}
\label{sec:exp_setup}

\paragraph{Settings.} We compare against a vanilla decoder-only Transformer and representative compute-decoupled baselines that scale capacity via token-indexed parameters or embedding modules (e.g., STEM, MoLE, and SCONE-style conditional memory) \citep{sadhukhan2026stemscalingtransformersembedding,mole_ref,scone_ref}. All methods share the same backbone at each scale (0.73B and 1.15B) and the same training token budget; differences are restricted to the token-indexed parameterization, feature extraction, and injection design. Backbone hyperparameters and training recipes are summarized in Table~\ref{tab:model_config} and Appendix~\ref{app:training_details} (extended setup in Appendix~\ref{app:exp_setup_full}).

\paragraph{Data.} We pretrain on \textsc{OLMo-mix-1124} \citep{olmo2_paper} (web + domain data from DCLM/Dolma) \citep{dclm_paper,dolma_paper} and evaluate with \texttt{lm-evaluation-harness} \citep{eval-harness} on a fixed downstream suite with consistent prompts and shot counts (Table~\ref{tab:eval_suite}).

\paragraph{Baseline.} For fairness, we keep backbone compute nearly unchanged across methods and report the additional activated compute/memory explicitly (Table~\ref{tab:inj_budgets}). Since lookup-based scaling is often limited by memory traffic, we align the memory budget $\mathcal{C}$ and the number of injected views/slots (or use fewer for \aname) and report the resulting budget accounting (Table~\ref{tab:budget_alignment}).

\paragraph{Main-table configurations.}
In Table~\ref{table:main-results}, the capacity multipliers 1$\times$, 2$\times$, and 4$\times$ refer to predefined combinations of inter-layer residual views ($h$) and attention-value views ($v$), rather than a single uniform table-scaling rule.
Because our backbones use grouped-query attention with $h_{\text{kv}}=h/2$, the injected value width is $d_{\text{kv}}=(h_{\text{kv}}/h)d=d/2$, so each $v$ table has half the width of an $h$ table.
Concretely, 1$\times$ uses $2v$ with kernels $\{3,5\}$, 2$\times$ uses $1h+2v$ with $h$ kernel $\{3\}$ and $v$ kernels $\{3,5\}$, and 4$\times$ uses $3h+2v$ with $h$ kernels $\{3,5,7\}$ and $v$ kernels $\{3,5\}$.
We denote the uncompressed setting by 100\% and the hybrid-hash setting by 50\% (compression ratio $\rho=0.5$); full configuration details are summarized in Appendix Table~\ref{tab:main_table_configs}.

\subsection{Result}
\begin{figure*}[!tp]  
\centering  
\includegraphics[width=\textwidth]{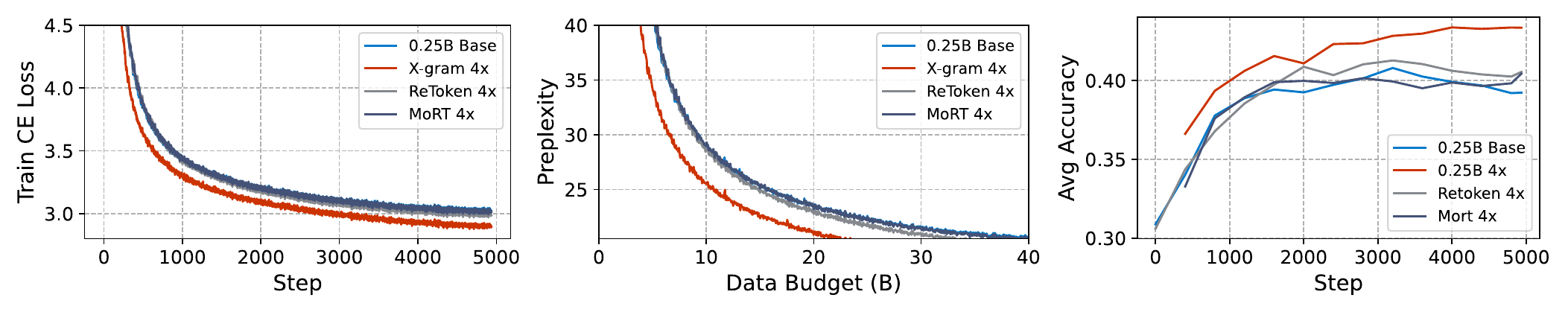}
\caption{
\textbf{Pretraining convergence with token-indexed injection.}
We report pretraining loss/PPL trajectories under representative injection configurations to illustrate the stability and efficiency of \aname-style designs.
}
\vspace{-2mm}
\label{fig:pt_loss}
\end{figure*}

\begin{table}[t]
\centering
\caption{Overall ablations of \aname. We report validation perplexity (PPL; lower is better) together with total parameters.}
\vspace{0.5mm}
\setlength{\tabcolsep}{3.5pt}
\footnotesize
\resizebox{0.7\columnwidth}{!}{%
\begin{tabular}{l|c|c|c}
\toprule
\textbf{Variant} & \textbf{PPL}$\downarrow$ & \textbf{Avg Acc}$\uparrow$ & \textbf{Total Params} \\
\midrule
\aname (full) & 17.702 & 49.68 & 0.73B + 4.67B \\
w/o depth-aware scaling \& warmup & 17.703 & 49.32 & 0.73B + 4.67B \\
w/o \aname Extraction & 18.420 & 47.80 & 0.73B + 4.67B \\
w/o Context-aware Gating & 17.920 & 48.41 & 0.73B + 4.67B \\
w/o Retrieved-sequence Normalization & 18.282 & 47.38 & 0.73B + 4.67B \\
w/o multi-view normalization & 17.707 & 48.74 & 0.73B + 4.67B \\
w/o LR scaling & 18.271 & 47.62 & 0.73B + 4.67B \\
\bottomrule
\end{tabular}%
}
\label{tab:abla_overall}
\vspace{-2mm}
\end{table}
\paragraph{Main Results and Downstream Evaluation.}
Table~\ref{table:main-results} reports zero/few-shot downstream accuracy under matched training budgets and comparable token-indexed parameter sizes.
Overall, \aname consistently outperforms strong retrieval baselines, and its advantage remains robust as capacity scales.
At 0.73B, \aname-50\% reaches 48.5 at 1$\times$, surpassing MoRT by +2.7 and Retoken by +3.5; the best \aname variant further reaches 49.7 at 2$\times$ and 49.5 at 4$\times$, yielding gains of +2.7 to +3.8 over MoRT and +1.3 to +3.9 over Engram.
At 1.15B, the best \aname variant reaches 49.5 at 1$\times$, 50.5 at 2$\times$, and 50.8 at 4$\times$, improving by +2.0 over Retoken at 1$\times$, +2.7 over MoRT at 2$\times$, and +2.3 over Engram at 4$\times$.
Crucially, the gains scale more reliably with memory: \aname continues to improve as capacity grows, whereas baseline scaling is less consistent, suggesting that additional memory is converted into usable signal rather than redundant capacity.

\paragraph{Training Stability and Data Efficiency.}
\aname also exhibits strong training stability and data efficiency (Figure~\ref{fig:pt_loss}).
Under the representative 4$\times$ setting, \aname maintains a persistent loss advantage of more than 0.04 over MoRT and Retoken from early to late training.
This stability translates into improved data efficiency: \aname reaches baseline-quality performance using only 57.25\% of the training data.
These results suggest that X-gram extracts reusable and information-dense patterns, turning the injected memory into effective capacity rather than brittle memorization.

\paragraph{Scalability and Parameter Efficiency.}
In terms of parameter efficiency, \aname benefits from frequency-aware hashing that compresses token-indexed memory without sacrificing quality. As shown in Table~\ref{tab:abla_hash}, Hybrid 1$\times$ (50\% memory) improves average accuracy from 47.53 to 48.48 while reducing the table parameters from 2.35B to 1.17B. Hybrid 2$\times$ further reaches 48.68 at the same 2.35B table-parameter budget as VIP-only, while also improving perplexity from 18.310 to 17.951, demonstrating higher utility per memory parameter.

\FloatBarrier

\subsection{Ablation Studies}
\label{sec:ablation}

We structure ablations into three compact groups that directly correspond to our motivating observations: (i) training stability and heterogeneous demand (gating and stabilization), (ii) long-tail trainability and parameter efficiency (hybrid hashing), and (iii) representation diversity under fixed slots (ShortConv design).

\paragraph{Overall ablations.}
Table~\ref{tab:abla_overall} supports the contribution of the main system components. Depth-aware scaling and warmup help stabilize training and improve final quality; the gating mechanism in ShortConv provides input-dependent feature selection that improves refinement quality; and stabilization together with sparse-aware optimization helps maintain stable optimization and improves final quality in the high-capacity setting.
\begin{figure*}[!tp]
\centering
\begin{subfigure}[t]{0.33\textwidth}
  \centering
  \includegraphics[width=\linewidth]{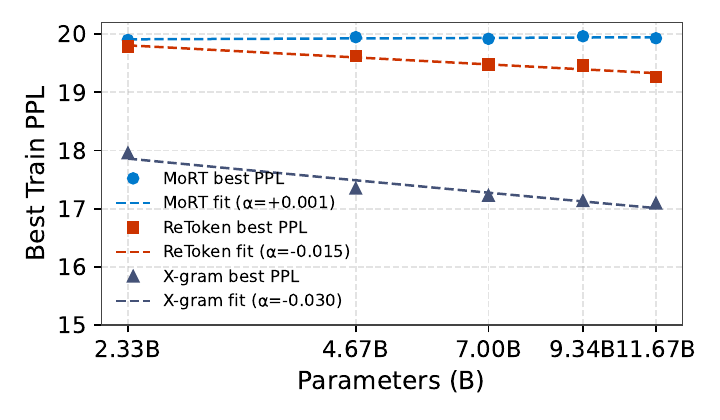}
\end{subfigure}%
\begin{subfigure}[t]{0.33\textwidth}
  \centering
  \includegraphics[width=\linewidth]{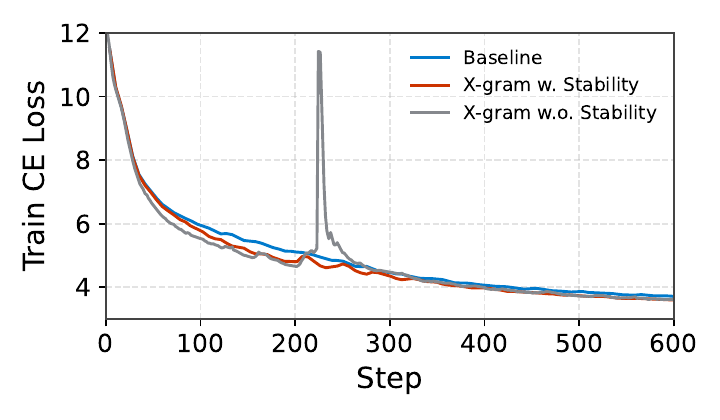}
\end{subfigure}%
\begin{subfigure}[t]{0.33\textwidth}
  \centering
  \includegraphics[width=\linewidth]{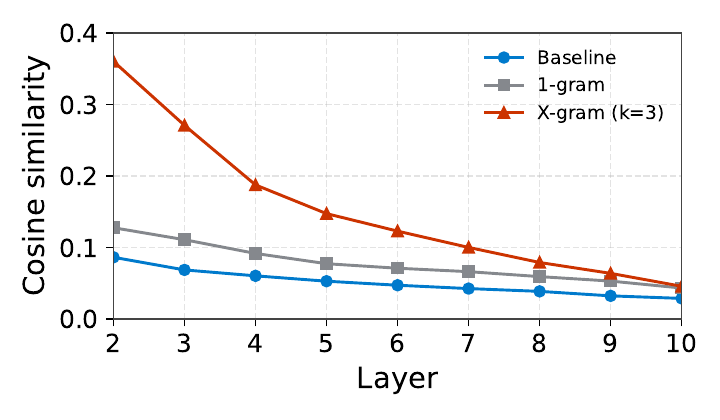}
\end{subfigure}
\caption{Analysis of injection trade-offs and representation behavior. Left: Pareto trade-offs under different injection targets, comparing quality versus parameter budgets. Middle: Training loss trajectories with and without warmup gating. Right: Cosine similarity between each layer's hidden state and the initial-layer hidden state, showing that \aname maintains stronger token identity signals across depth.}
\label{fig:qkvo_pareto}
\vspace{-2mm}
\end{figure*}

\paragraph{Hybrid hashing ablations.} Table~\ref{tab:abla_hash} isolates the effect of frequency-aware allocation. VIP-only (no hashing) is the strongest contrast but is substantially more parameter-expensive, whereas hybrid hashing concentrates capacity where gradients are dense and amortizes the long tail into shared buckets. The ablations on row-wise gating and alias copies suggest two complementary roles: suppressing collision noise under aggressive compression, and recycling unused capacity without exploding the parameter footprint.
Appendix~\ref{app:hash_diagnostics} reports the corresponding row-level statistics: relative to no hashing, frequency-aware routing lowers hotspot concentration and sharply reduces dispersion in row-level movement relative to initialization.

\paragraph{ShortConv design ablations.}
Table~\ref{tab:abla_shortconv} examines how temporal refinement alleviates fixed-slot collapse. In particular, multi-scale extraction introduces local structure beyond static token identity, and the results suggest that combining multi-scale refinement with view-specific kernels yields the best performance among the tested configurations, with only modest extra compute relative to the backbone.

\begin{table*}[t]
\centering
\newlength{\ablationcaptionheight}
\setlength{\ablationcaptionheight}{29mm}
\begin{minipage}[t]{0.485\textwidth}
\centering
\begin{minipage}[t][\ablationcaptionheight][t]{\linewidth}
\captionof{table}{Ablations on frequency-aware memory allocation. We compare VIP-only routing and hybrid routing variants under matched model scale and report validation perplexity and downstream average accuracy.}
\label{tab:abla_hash}
\end{minipage}
\vspace{0.5mm}
\setlength{\tabcolsep}{3.5pt}
\footnotesize
\resizebox{\linewidth}{!}{%
\begin{tabular}{l|c|c|c}
\toprule
\textbf{Routing / Allocation} & \textbf{Table Params} & \textbf{PPL}$\downarrow$ & \textbf{Avg Acc.} \\
\midrule
VIP-only (no hash) 1x & 0.73B + 2.35B & 18.310 & 47.53 \\
Hybrid 1x (50\% memory) & 0.73B + 1.17B & 18.419 & 48.48 \\
Hybrid 1x w/o gating $a_{\ell,j}$ & 0.73B + 1.17B & 18.415 & 47.47 \\
Hybrid 1x w/o alias copies ($K{=}1$) & 0.73B + 1.17B & 18.337 & 48.24 \\
Hybrid 2x (50\% memory) & 0.73B + 2.35B & 17.951 & 48.68 \\
\bottomrule
\end{tabular}%
}
\end{minipage}
\hfill
\begin{minipage}[t]{0.485\textwidth}
\centering
\begin{minipage}[t][\ablationcaptionheight][t]{\linewidth}
\captionof{table}{ShortConv design ablations. ShortConv does not materially change the token-indexed table size; we therefore report relative FLOPs per token for the refinement module.}
\label{tab:abla_shortconv}
\end{minipage}
\vspace{0.5mm}
\setlength{\tabcolsep}{3.5pt}
\footnotesize
\resizebox{\linewidth}{!}{%
\begin{tabular}{l|c|c|c}
\toprule
\textbf{ShortConv Variant} & \textbf{PPL}$\downarrow$ & \textbf{Avg Acc.}$\uparrow$ & \textbf{FLOPs/token} \\
\midrule
Single-kernel ($k{=}3$), shared & 18.031 & 49.18 & $1.01\times$ \\
Multi-scale ($k{\in}\{2,3,4,5\}$) & 18.014 & 48.46 & $1.02\times$ \\
Multi-scale ($k{\in}\{2,4,6,8\}$) & 17.958 & 48.91 & $1.02\times$ \\
Multi-scale ($k{\in}\{1,3,5,7\}$) & 18.054 & 48.86 & $1.02\times$ \\
Multi-scale, $k{\in}\{3,5,7,9\}$, (ours) & 17.924 & 49.27 & $1.02\times$ \\
Multi-scale ($k{\in}\{3,6,9,12\}$) & 17.941 & 49.14 & $1.02\times$  \\
\bottomrule
\end{tabular}%
}
\end{minipage}
\vspace{1mm}
\end{table*}

\FloatBarrier

\subsection{Analysis}

\paragraph{Why is \aname better?}
Large LMs often rely on depth to gradually compose features that approximate a knowledge lookup; an explicit retrieval channel can therefore act like increasing the \emph{effective depth} by reducing early-stage feature-composition burden.
To test whether \aname injects richer signals (and why it improves so strongly over the baseline), we compare (i) a baseline model, (ii) direct 1-gram embedding injection, and (iii) \aname's X-gram injection (1-gram retrieval refined by ShortConv with a representative $k{=}3$ kernel).
We measure cosine similarity between each layer's hidden state and the initial-layer hidden state.
As shown in the right panel of Figure~\ref{fig:qkvo_pareto}, similarity decreases with depth for all methods, but \aname maintains substantially higher cosine similarity than direct 1-gram injection from early to late layers, suggesting that X-gram refinement provides more informative, depth-consistent signals that make representations prediction-ready earlier.

\begin{figure}[!tbp]
\centering
\includegraphics[width=0.6\columnwidth]{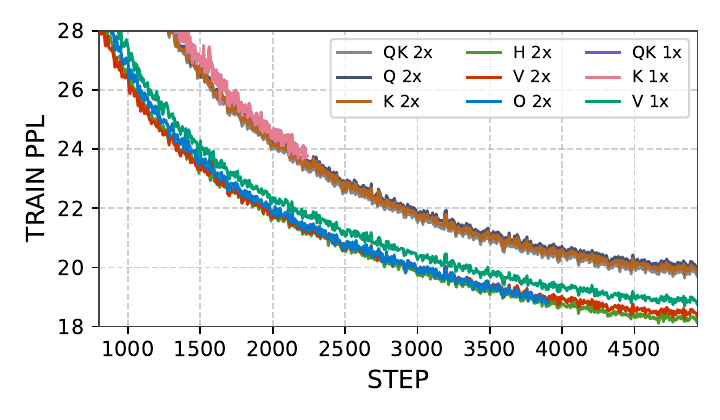}
\caption{Depth-wise sensitivity of injection sites. We vary the injection location (inter-layer residual vs Q/K/V/O) across layers and report the resulting PPL.}
\label{fig:inject_depth_site}
\vspace{1mm}
\end{figure}

\paragraph{Where to inject.}
We further ablate the \emph{injection placement} to identify the most effective sites under a fixed budget. Concretely, we compare inter-layer residual injection and attention-stream injection to $\mathbf{Q}$, $\mathbf{K}$, $\mathbf{V}$, and $\mathbf{O}$ (Section~\ref{sec:method:inject}).Figure~\ref{fig:inject_depth_site} shows the resulting training perplexity trajectories under these choices. Overall, injecting into the value stream or the inter-layer residual path yields the most favorable trade-offs, while injection into $\mathbf{Q}$, $\mathbf{K}$, or their combination is consistently less effective in our setting. These results motivate our focus on value-stream and inter-layer residual injection in the main design.

\begin{figure*}[!tp]
\centering
\begin{subfigure}[t]{0.48\textwidth}
  \centering
  \includegraphics[width=\linewidth]{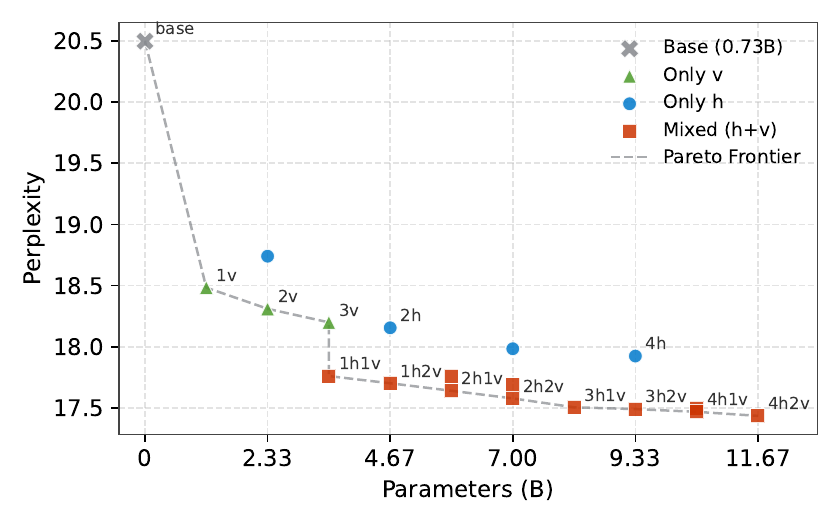}
\end{subfigure}%
\hfill
\begin{subfigure}[t]{0.48\textwidth}
  \centering
  \includegraphics[width=\linewidth]{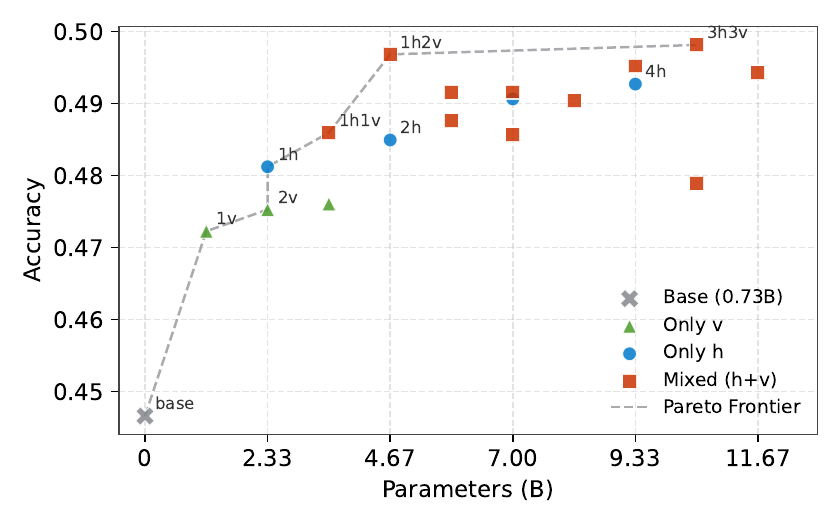}
\end{subfigure}
\caption{Pareto frontier analysis of injection pathways. We compare attention-value views (v), inter-layer residual views (h), and mixed configurations (h+v) under varying parameter budgets in terms of perplexity (left) and downstream accuracy (right). Mixed injection forms the best overall Pareto frontier in our tested configurations.}
\label{fig:hv_pareto}
\vspace{-2mm}
\end{figure*}

To further quantify the complementarity between injection pathways, we evaluate a range of representative combinations of inter-layer residual views (h) and attention-value views (v) at the 0.73B scale, reporting accuracy and perplexity as a function of total injected parameters (Figure~\ref{fig:hv_pareto}). Value-only injection yields strong gains at small parameter budgets but tends to plateau as capacity increases. Inter-layer-only injection remains competitive at larger budgets, but mixed (h+v) configurations dominate the Pareto frontier on both metrics. This suggests that the two pathways are complementary and that combining them provides a better quality–efficiency trade-off than using either pathway alone.

\begin{figure*}[!tp]
\centering
\begin{subfigure}[t]{0.48\textwidth}
  \centering
  \includegraphics[width=\linewidth]{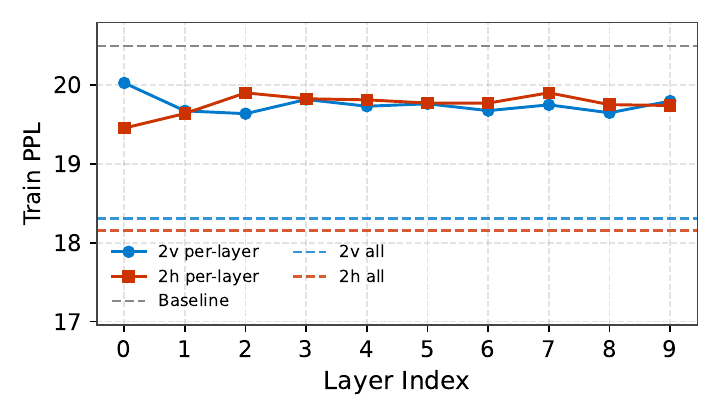}
\end{subfigure}%
\hfill
\begin{subfigure}[t]{0.48\textwidth}
  \centering
  \includegraphics[width=\linewidth]{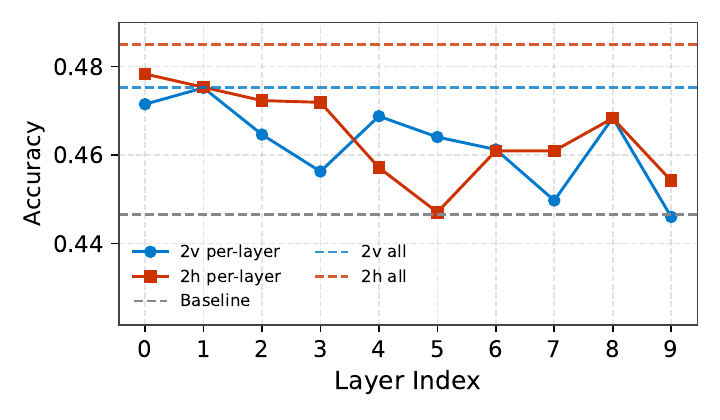}
\end{subfigure}
\caption{Per-layer injection ablation. Each point shows the result of injecting at only one layer (2v / 2h per-layer) compared with full-layer injection (2v / 2h all) in terms of training perplexity (left) and accuracy (right). Single-layer injection consistently outperforms the baseline, with early layers showing higher accuracy sensitivity, yet full-layer injection remains significantly superior.}
\label{fig:layer_ablation}
\vspace{-2mm}
\end{figure*}

\paragraph{Depth-wise sensitivity.}
To further investigate the role of memory injection across network depth, we perform a per-layer ablation for the 2v and 2h strategies. We restrict injection to a single layer (Layer 0 through 9) and compare against full-layer injection (all layers) and the baseline (Figure~\ref{fig:layer_ablation}). Injecting at any single layer consistently outperforms the baseline, indicating that external memory signals are useful throughout the network. In terms of accuracy, earlier layers tend to yield larger gains than mid-to-late layers, suggesting that shallow layers are more sensitive to the injected memory signal. However, regardless of which layer is chosen for single-layer injection, a substantial gap remains relative to full-layer injection in both perplexity and accuracy. This indicates that the effectiveness of \aname\ is not explained by added parameters alone, but also by distributing the injected signal across depth so that multiple layers can repeatedly incorporate the retrieved features.

\paragraph{Row-level Utilization under Frequency-aware Hashing}
\label{app:hash_diagnostics}
We further characterize how frequency-aware hashing changes the utilization of physical rows under compression.
We consider two row-level statistics.
First, we sort physical rows by hit count in descending order and report the fraction of total hits absorbed by the top 0.1\%, 1\%, and 10\% rows.
Second, for each injected table, we compute the normalized movement of every reachable physical row relative to initialization,
\begin{equation}
m_j=\frac{\|\mathbf{B}[j]-\mathbf{B}^{(0)}[j]\|_2}{\|\mathbf{B}^{(0)}[j]\|_2},
\end{equation}
and summarize within-table dispersion using both the coefficient of variation $\mathrm{CV}=\mathrm{std}(m_j)/\mathrm{mean}(m_j)$ and a strong/weak row movement gap defined as the ratio between the p75 and p25 values of $\{m_j\}$.
For the movement-based quantities, Table~\ref{tab:hash_diagnostics} reports median values across injected tables.

\begin{table}[t]
\centering
\caption{Row-level access concentration and trainability under different routing schemes. The top block reports the fraction of total hits absorbed by the hottest physical rows. The bottom block reports normalized row movement relative to initialization; values are median across injected tables.}
\label{tab:hash_diagnostics}
\vspace{0.5mm}
\footnotesize
\setlength{\tabcolsep}{4pt}
\begin{tabularx}{0.7\columnwidth}{@{}>{\RaggedRight\arraybackslash}Xcc@{}}
\toprule
\textbf{Metric} & \textbf{No Hash} & \textbf{Freq.-aware Hash} \\
\midrule
\multicolumn{3}{@{}l}{\textbf{Row-hit concentration}} \\
Top 0.1\% rows hit ratio & 57.74\% & 43.11\% \\
Top 1\% rows hit ratio & 76.10\% & 66.31\% \\
Top 10\% rows hit ratio & 93.66\% & 86.77\% \\
\midrule
\multicolumn{3}{@{}l}{\textbf{Row-level trainability (normalized L2 movement)}} \\
Global CV (std / mean) & 0.587 & 0.159 \\
Strong/weak row movement gap & $3.47\times$ & $1.14\times$ \\
\bottomrule
\end{tabularx}
\vspace{-1mm}
\end{table}

The no-hash lookup is substantially more head-concentrated: a small prefix of rows absorbs most accesses, whereas frequency-aware hashing noticeably flattens the access distribution.
Row-level training is also much more uneven without hashing: the within-table CV reaches 0.587 and the p75/p25 movement gap is $3.47\times$, whereas frequency-aware hashing reduces them to 0.159 and $1.14\times$, respectively.
Because these movement metrics are defined as within-table relative quantities, they are not directly driven by absolute table size.
Overall, frequency-aware hashing does not make row usage uniform, but under a fixed budget it reduces hotspot domination and makes row-level training substantially more balanced.

\FloatBarrier



\subsection{Related Work}
\label{app:related_work_full}

\paragraph{Lookup-based scaling and conditional memory.}
A recent thread explores adding large token- or $n$-gram-indexed parameters as an auxiliary memory channel, aiming to scale capacity with minimal additional FLOPs.
Mixture-of-lookup designs route tokens to a small set of retrieved vectors and combine them with lightweight mixing, enabling substantial parameter growth while keeping the runtime compute largely unchanged \citep{mole_ref}.
Similarly, scalable lookup-based conditional memory frames retrieval as a new axis of sparsity: additional capacity primarily increases memory traffic rather than matrix multiplication \citep{scone_ref}.
These methods are attractive in practice because retrieval is deterministic from token IDs and can be served from host memory with prefetching, but they also expose a core bottleneck: the efficiency of \emph{using} those parameters per accessed byte, especially under Zipfian long tails where many rows are rarely updated.

\paragraph{Embedding and vocabulary scaling.}
Another line of work enlarges the input interface by scaling embedding layers, vocabulary size, or related embedding modules \citep{yu2025scalingembeddinglayerslanguage, huang2025overtokenizedtransformervocabularygenerally, sadhukhan2026stemscalingtransformersembedding}.
Closely related work also studies memory-efficient embedding parameterizations through heterogeneous capacity allocation, compositional compression, and parameter sharing, especially in recommendation systems and large-vocabulary language models \citep{naumov2019deep, ginart2021mixed, shi2020compositional, desai2022random, desai2022trade, desai2023defense, baevski2018adaptive, svenstrup2017hash}.
These approaches mainly improve the capacity or memory efficiency of input-side embeddings, often through non-uniform allocation or shared parameterization.
In contrast, we study layer-wise token-indexed memory rather than a single input embedding layer, and use frequency information to structure physical capacity allocation, with hashing applied only locally within dense buckets where collisions are unavoidable.

\paragraph{Sparse MoE Models.} Sparse Mixture-of-Experts (MoE) \citep{shazeer2017outrageously} employs learnable gating to activate a sparse subset of experts per token, enabling efficient capacity scaling \citep{lepikhin2020gshard, fedus2022switch, jiang2024mixtral}. While recent advances explore fine-grained specialization \citep{dai2024deepseekmoe} and unified scaling laws \citep{krajewski2024scaling}, standard MoE designs typically partition experts by layer. This layer-wise isolation causes total memory to scale linearly with depth and limits the potential for cross-layer operator reuse. Existing routing strategies—ranging from load-balancing heuristics \citep{fedus2022switch, lepikhin2020gshard} and hash-based assignments \citep{roller2021hash} to expert-choice \citep{zhou2022mixture}, BASE layers \citep{lewis2021base}, and soft mixtures \citep{puigcerver2024from}—primarily optimize intra-layer token allocation. However, these methods do not address the recursive, multi-layer exposure of experts, which is central to the universal expert design in \aname.

\paragraph{Injection sites and value-path augmentation.}
Where injection occurs matters for both stability and efficiency.
Injecting into the attention value pathway can be particularly cost-effective under grouped-query attention because the value dimensionality is smaller than the query/key dimensionality, yielding lower activation and compute overhead for the same memory budget.
Recent analyses of value-residual designs further highlight the importance of value-stream information flow \citep{zhou2025valueresiduallearning}.
Our method adopts value-path injection as the default site and uses depth-aware gating and normalization to keep the injected signal stable during training.

\paragraph{Positioning of \aname.}
\aname is motivated by a gap in prior compute-decoupled baselines: while MoLE/SCONE/STEM demonstrate that token-indexed capacity \emph{can} be scaled, they do not systematically analyze the \emph{training dynamics} of token-indexed parameters under Zipfian data, nor do they quantify \emph{where} (which layers, which components) injection is actually demanded.
As a result, their added parameters are often activated in a coarse, uniform manner, which (i) wastes updates on under-trained long-tail rows, and (ii) encourages redundant, low-distinctiveness slot representations, limiting the amount of information extracted per accessed byte.
In contrast, \aname explicitly links design choices to observed failure modes: frequency-aware allocation improves trainability under long-tail sparsity, depth/component-aware injection targets demand heterogeneity, and ShortConv promotes diversity to avoid fixed-slot collapse.

\section{Conclusion}

We analyzed why naive lookup-based scaling saturates and identified two root causes: long-tail under-training and representational collapse across fixed slots. To address them, \aname introduces probability-balanced hybrid hashing to improve long-tail trainability, a lightweight ShortConv to extract multi-scale x-gram features and avoid redundant slots, and depth-aware injection that aligns static memory with attention content pathways. Across 0.73B and 1.15B models, \aname consistently improves downstream averages at matched budgets, with gains up to 3--4 points over strong baselines while using smaller tables (Table~\ref{table:main-results}). These results indicate that token-indexed memory can be a practical scaling axis when it is trainable, diverse, and demand-aware.

\newpage
\bibliography{ref}
\newpage
\appendix

\section{Acknowledge}

We would like to sincerely thank Jiacheng You and Songling Yang for their insightful discussions and valuable suggestions that greatly improved this work. We also extend our deep appreciation to all researchers—both publicly and anonymously involved in discussions and the review process—for their time, effort, and thoughtful feedback, which have been instrumental in refining this study.

\section{Notation}
\label{app:notation}
\begin{table*}[h]
\centering
\caption{Summary of mathematical notation used throughout the paper.}
\label{tab:notation}
\vspace{0.5mm}
\small
\setlength{\tabcolsep}{5pt}
\begin{tabularx}{\textwidth}{@{}l X@{}}
\toprule
\textbf{Symbol} & \textbf{Meaning} \\
\midrule
$\tau$ & Tokenizer. \\
$\mathcal{V}$, $|\mathcal{V}|$ & Vocabulary and vocabulary size. \\
$\mathbf{x}=(x_1,\ldots,x_T)$ & Tokenized input sequence of length $T$. \\
$\mathbf{E}$ & Input token embedding matrix. \\
$\ell$, $L$ & Transformer layer index and total number of layers. \\
$\mathbf{h}_{\ell,t}$, $\mathbf{h}_t$, $\mathbf{H}_{\ell}$ & Hidden state at layer $\ell$ and position $t$, position-wise hidden state when the layer is clear, and the full sequence of hidden states. \\
$\mathbf{y}_t$ & Feed-forward block output before token-indexed modulation. \\
$d$, $d_s$, $d_{\mathrm{kv}}$ & Backbone hidden width, injected feature width, and key/value width. \\
$m$, $M_{\ell}$ & Injection view index and number of views at layer $\ell$. \\
$s$ & Injection site, such as attention value stream or inter-layer residual stream. \\
$\mathbf{B}_{\ell}^{(m)}$ & Lookup table for layer $\ell$ and view $m$. \\
$S$, $\rho$ & Number of physical lookup rows and compression ratio. \\
$\mathcal{R}_{\phi}$ & Token-to-parameter routing operator. \\
$\mathcal{T}_{\psi}$ & Information extraction operator. \\
$\mathcal{I}_{s}$ & Injection operator at site $s$. \\
$\tilde{\mathbf{E}}_{\ell}^{(m)}$ & Extracted feature sequence from view $m$. \\
$\Delta_{\ell}$, $\Delta_{\ell,t}$ & Fused injection signal for layer $\ell$ and position $t$. \\
$g_{\ell}^{(m)}$ & Layer-wise or view-wise injection gate. \\
$g_t$ & Engram context gate computed from the RMS-normalized key-query score with the sign-preserving square-root transform before the sigmoid. \\
$\mathbf{s}_{\ell}$ & Layer-wise scaling vector. \\
$\mathbf{W}_r$, $\mathbf{W}_v$, $\mathbf{W}_k$ & Router, value projection, and key projection matrices. \\
$\mathcal{C}$, $\mathcal{A}$, $\mathcal{F}$ & Memory, activation, and compute budgets. \\
$\mathcal{A}(\omega)$ & Physical lookup rows accessed by token $\omega$. \\
$c_{\omega,j}$, $a_{\ell,j}^{(m)}$ & Aggregation coefficient and row-wise learnable scale for physical row $j$. \\
$K$ & Number of retrieved rows per token in \aname routing. \\
$k$ & ShortConv kernel size. \\
$\mathcal{N}$, $h_n(\cdot)$ & Set of explicit $n$-gram orders and the corresponding hash function. \\
$S_n$, $d_n$, $H_n$ & Table size, feature width, and number of hash heads for an $n$-gram table. \\
$d_e$ & Concatenated Engram feature width. \\
$k_{\mathrm{top}}$ & Sparse top-$k$ router size in MoRT. \\
$f_{\theta}$, $f_{\theta,\varphi}$ & Backbone model and lookup-augmented model. \\
$\mathcal{D}$, $\ell(\cdot)$ & Training distribution and language-modeling loss. \\
\bottomrule
\end{tabularx}
\vspace{-2mm}
\end{table*}

\section{Full Versions of Compressed Sections}
\label{app:full_versions}

\subsection{Extended Experimental Setup}
\label{app:exp_setup_full}
We provide additional experimental details that are compressed in Section~\ref{sec:exp_setup}.
\textbf{Baselines.}
We compare against (i) \textbf{Vanilla Transformer}, the standard decoder-only backbone, and representative compute-decoupled baselines that scale capacity via embedding modules or token-indexed memory, including \textbf{STEM} \citep{sadhukhan2026stemscalingtransformersembedding}, \textbf{MoLE} \citep{mole_ref}, and \textbf{SCONE} \citep{scone_ref}.
All models share identical backbones (same depth/width) and are trained under the same token budget; methods differ only in the token-indexed parameterization and injection design.
\textbf{Data and evaluation.}
Models are pretrained on \textbf{\textsc{OLMo-mix-1124}} \citep{olmo2_paper}, which integrates high-quality web text from \textbf{DCLM} \citep{dclm_paper} with diverse domain-specific data from \textbf{Dolma} \citep{dolma_paper}.
We evaluate using \texttt{lm-evaluation-harness} \citep{eval-harness} and keep prompts/shot counts fixed across methods within each benchmark (Table~\ref{tab:eval_suite}).
\textbf{Budget alignment.}
We keep additional compute from lookup and ShortConv within a small constant factor (Table~\ref{tab:inj_budgets}) and align the memory budget $\mathcal{C}$ and injected views/slots across token-indexed methods (Table~\ref{tab:budget_alignment}).

\subsection{Training Details}\label{app:training_details}
We summarize the training configuration referenced in Section~\ref{sec:exp_setup}.
Unless otherwise stated, all models share the same backbone architecture, tokenizer, data mixture, and total training tokens; we only vary the token-indexed parameterization.

\paragraph{Backbone configurations.}
Table~\ref{tab:model_config} lists the backbone hyperparameters used at each scale.

\begin{table}[t]
\centering
\caption{Backbone configurations used in our experiments. We report the architectural hyperparameters that fully determine the compute profile of the Transformer blocks.}
\vspace{0.5mm}
\setlength{\tabcolsep}{3.5pt}
\footnotesize
\resizebox{0.5\columnwidth}{!}{%
\begin{tabular}{l|cc}
\toprule
\textbf{Backbone setting} & \textbf{\textsc{Small}} & \textbf{\textsc{Medium}} \\
\midrule
Layers ($L$) & 10 & 12 \\
Model width ($d$) & 1536 & 2048 \\
FFN width ($d_{\mathrm{ff}}$) & 4096 & 5120 \\
Attention heads ($h$) & 12 & 16 \\
KV heads ($h_{\mathrm{kv}}$) & 6 & 8 \\
Sequence length & 8192 & 8192 \\
Vocabulary size & 151{,}936 & 151{,}936 \\
\bottomrule
\end{tabular}%
}
\label{tab:model_config}
\vspace{-2mm}
\end{table}

\paragraph{Optimization hyperparameters.}
We use AdamW with a cosine decay schedule and warmup, gradient clipping, and mixed precision; Table~\ref{tab:training_details} lists the key hyperparameters.
Unless otherwise stated, these settings are identical across baselines and \aname.
\begin{table}[t]
\centering
\caption{Training hyperparameters. Unless stated otherwise, we use the same recipe for baselines and \aname, and only change the token-indexed parameterization (Table~\ref{tab:budget_alignment}).}
\vspace{0.5mm}
\setlength{\tabcolsep}{3.5pt}
\footnotesize
\resizebox{0.8\columnwidth}{!}{%
\begin{tabular}{l|c}
\toprule
\textbf{Item} & \textbf{Value} \\
\midrule
Optimizer & AdamW \\
Adam $\beta$ / $\epsilon$ & $(0.9, 0.95)$ / $10^{-8}$ \\
Weight decay & $0.1$ \\
Gradient clipping & $1.0$ (global norm) \\
LR schedule & cosine with warmup \\
Warmup & $5\%$ of total steps \\
Peak LR & $1\mathrm{e}{-3}$ (\textsc{Small}); $7\mathrm{e}{-4}$ (\textsc{Medium}) \\
Precision & BF16 params, FP32 reductions \\
Activation checkpointing & full (per block) \\
Sequence length & $8192$ \\
Global batch (samples) & $1024$ \\
Micro-batch (per GPU) & $8$ \\
Gradient accumulation & $8$ \\
Total train tokens & $41$B \\
Total steps (approx.) & $4.9$k \\
Parallelism & HSDP 2D mesh (dp\_replicate=2, dp\_shard=8) \\
Compile & \texttt{torch.compile} enabled \\
Streaming chunk size & $40960$ tokens \\
Streaming prefetch & queue size $16$, \texttt{native:truncate} pack \\
Tokenizer & Qwen3-0.6B \\
\bottomrule
\end{tabular}%
}
\label{tab:training_details}
\vspace{-2mm}
\end{table}

\paragraph{Hardware and schedule.}
Unless otherwise stated, runs are executed with distributed BF16 training using NCCL and HSDP on available GPUs. Total training tokens are 41B (about 4.9k steps at the default batch/sequence settings).

\paragraph{Budget alignment details.}
For token-indexed methods, we explicitly report the memory budget $\mathcal{C}$ and keep the injected compute within a small constant factor of the backbone.
When comparing lookup-based approaches, we match the number of injected slots/views (or use fewer for \aname) and compare under the same or smaller $\mathcal{C}$ by construction (Table~\ref{tab:budget_alignment}).
\begin{table}[t]
\centering
\caption{Budget alignment and fairness protocol. We align methods under nearly identical backbone compute, and compare token-indexed approaches under the same (or fewer) active slots while reporting memory cost $\mathcal{C}$ explicitly. Values are reported per injected layer and per view unless noted.}
\vspace{0.5mm}
\setlength{\tabcolsep}{3.5pt}
\footnotesize
\resizebox{0.8\columnwidth}{!}{%
\begin{tabular}{l|c|c|c}
\toprule
\textbf{Method} & \textbf{\# slots/views} & $\Delta\mathcal{F}$ (\textbf{rel.}) & $\mathcal{C}$ (\textbf{dominant}) \\
\midrule
Vanilla (no lookup) & 0 & $1.00\times$ & 0 \\
Lookup baseline (uniform table) & $M$ & $\approx 1.00$--$1.02\times$ & $|\mathcal{V}|\,d_s$ \\
\aname (VIP+hash+ShortConv) & $\le M$ & $\approx 1.00$--$1.02\times$ & $\rho|\mathcal{V}|\,d_s$ \\
\bottomrule
\end{tabular}%
}
\label{tab:budget_alignment}
\vspace{-2mm}
\end{table}

\paragraph{Evaluation protocol.}
We evaluate using a unified harness configuration, keeping prompts and shot counts fixed across methods within each benchmark.
We report mean and standard error over multiple seeds where applicable.
\begin{table}[t]
\centering
\caption{Downstream evaluation suite and shot settings. We use the same prompts and shot counts for all methods within each benchmark.}
\vspace{0.5mm}
\setlength{\tabcolsep}{3.5pt}
\footnotesize
\resizebox{0.5\columnwidth}{!}{%
\begin{tabular}{l|c|c}
\toprule
\textbf{Benchmark} & \textbf{Category} & \textbf{Shots} \\
\midrule
SciQ & reasoning & 0 \\
PIQA & reasoning & 0 \\
WinoGrande & reasoning & 0 \\
HellaSwag & reasoning & 0 \\
ARC-Easy & reasoning & 0 \\
ARC-Challenge & reasoning & 5 \\
\midrule
BoolQ & understanding & 0 \\
CommonsenseQA & understanding & 0/5 \\
OpenBookQA & understanding & 0 \\
SocialIQA & understanding & 0 \\
\midrule
MMLU & knowledge & 5 \\
\bottomrule
\end{tabular}%
}
\label{tab:eval_suite}
\vspace{-2mm}
\end{table}


\paragraph{Main-table configuration mapping.}
Table~\ref{tab:main_table_configs} summarizes the exact correspondence between the 1$\times$, 2$\times$, and 4$\times$ multipliers in Table~\ref{table:main-results} and their underlying $h/v$ view allocations, kernel choices, and compression settings.
\begin{table*}[!t]
\centering
\caption{Configuration mapping for the capacity multipliers used in Table~\ref{table:main-results}. Under the GQA backbones used in our experiments, $d_{\text{kv}}=(h_{\text{kv}}/h)d=d/2$, so each attention-value table ($v$) has half the width of an inter-layer residual table ($h$). We denote the uncompressed setting by 100\% and the hybrid-hash setting by 50\% (compression ratio $\rho=0.5$).}
\vspace{0.5mm}
\setlength{\tabcolsep}{5pt}
\footnotesize
\begin{tabular}{c|c|c|c}
\toprule
\textbf{Multiplier} & \textbf{View Configuration} & \textbf{$h$ Kernels} & \textbf{$v$ Kernels} \\
\midrule
1$\times$ & $2v$ & -- & $\{3,5\}$ \\
2$\times$ & $1h+2v$ & $\{3\}$ & $\{3,5\}$ \\
4$\times$ & $3h+2v$ & $\{3,5,7\}$ & $\{3,5\}$ \\
\bottomrule
\end{tabular}
\label{tab:main_table_configs}
\vspace{-2mm}
\end{table*}

\subsection{Observation (extended)}
\label{app:full_observation}
We conduct all observations by training two models under identical data and compute budgets: a \textbf{Base} Transformer and a \textbf{PLE} variant that injects token-indexed memory into every layer.
In the PLE model, each layer aggregates $M_\ell$ token-indexed views and injects their normalized sum into the layer according to the PLE pipeline described in Section~\ref{sec:method}.
Concretely, we use Eq.~(\ref{eq:obs_inj}), and we report statistics after training for 41B tokens.
We provide exact definitions for the activation norm and similarity metrics in Appendix~\ref{appobs:metrics}.

\textbf{Trainability is dominated by Zipfian frequency.}
To sharpen the optimization target, we first ask a basic but overlooked question: when a token-indexed lookup table is scaled up, are its parameters actually trainable under realistic data budgets?
In practice, the answer is constrained by Zipf's law.
Because token and $n$-gram sequences follow an extreme long tail, updates to large tables become highly sparse: most entries are rarely, if ever, visited, and therefore remain effectively under-trained.
This under-training is not a minor edge case but a structural property of frequency distributions.
For example, after processing 100M tokens, only $7.3\%$ of entries in a 2M-entry table receive more than 100 updates, whereas a 32K table reaches $97.6\%$ under the same criterion.
To quantify utilization, we compute the average activation magnitude for each token as the mean $L_2$ norm of its injected vector over all occurrences and layers (Appendix~\ref{appobs:metrics}).

\textbf{Injection demand varies across depth and modules.}
\label{app:obs_need}
We next ask where token-indexed injection is actually needed.
We measure, for each layer, how similar the layer's hidden representations are to the corresponding token embeddings at the input, and we further separate the analysis by module outputs.
This reveals a consistent depth-dependent drift: similarity decays with depth, indicating that token identity becomes progressively diluted as representations are repeatedly mixed.
Moreover, injection demand is heterogeneous across modules, with a clear preference for modulating attention content pathways and a stronger need for injection in deeper layers.
We compute similarity using cosine alignment between layer outputs and the input token embedding, averaged over tokens; the exact formula is given in Appendix~\ref{appobs:metrics}.

\textbf{Fixed-slot memory collapses to redundant subspaces.}
Finally, we analyze why naive fixed-slot designs often saturate quickly.
When multiple parallel lookup tables are used as fixed ``slots'' per layer, each slot is trained on highly overlapping token occurrences and receives similar gradients.
As a consequence, the learned representations often collapse to near-indistinguishable embeddings rather than specializing.
To diagnose this collapse, we compute pairwise cosine similarity between the retrieved embeddings across slots.
This lack of diversity reduces the effective information capacity per slot, so adding more tables primarily increases storage while yielding diminishing performance returns.

\section{Additional Details}

\subsection{Injection Sites and Efficiency (extended)}
\label{app:inj_sites_eff_full}
This appendix complements the concise discussion with additional intuition about why injection location is a first-order efficiency choice.
Injection determines not only \emph{where} the lookup signal enters the network, but also how budgets are spent and how much of the signal can be contextually matched.

\paragraph{Budgets and dominant terms.}
Under a fixed training recipe, lookup-based methods are typically dominated by memory budget $\mathcal{C}$: scaling the number of physical rows $S$ quickly dwarfs other parameter terms.
At the same time, the per-token activation and compute overhead are largely controlled by the injected dimensionality $d_s$, making injection location a lever for quality-per-budget (Table~\ref{tab:inj_budgets}).

\paragraph{Why attention content injection is efficient.}
Injecting into attention exposes the lookup signal to contextual matching: attention weights can select, suppress, or combine injected content based on the current hidden state.
Among attention pathways, value-stream injection is often the best trade-off because it augments content without directly perturbing similarity scores.
Under grouped-query attention, $d_{\text{kv}}$ is typically smaller than $d$, which further reduces the activation footprint relative to sites that require full-width injection.
By contrast, injecting into the score path (Q/K) can be substantially more expensive and more fragile, because it perturbs the matching mechanism itself and can dominate early optimization if not carefully warmed up and scaled.

\paragraph{Inter-layer residual injection as an anchor.}
An inter-layer residual path provides a complementary role: it supplies a stable additive signal that is less sensitive to the internal attention/MLP parameterization.
This is particularly useful when the long tail is aggressively compressed and collisions increase noise in the retrieved vectors.
In our design, we treat this path as optional stabilization rather than the primary interface for contextual matching.

\paragraph{Connection to \aname.}
These considerations motivate focusing on value-stream injection as the main interface between static memory and contextual computation, while using an inter-layer residual path when stability is the limiting factor.
Section~\ref{sec:method:inject} provides the explicit injection equations, and Section~\ref{sec:ablation} validates the resulting Pareto behavior across injection targets.

\subsection{Additional Details of Frequency-Aware Hash Mapping}
\label{app:routing_full}
This section provides additional construction details and experimental settings for the frequency-aware hash mapping introduced in the main text. We first estimate the empirical frequency of each token from a sampled corpus and denote it by $p(\omega)$. We then define a smoothed mass
\begingroup

\begin{equation*}
s(\omega)=p(\omega)^{\alpha}, \qquad 0<\alpha\le 1.
\end{equation*}
\endgroup

We first extract a small set of the most frequent tokens as the head set $\mathcal{V}_{\mathrm{vip}}$, and assign dedicated physical rows to them in a reserved VIP region. The remaining tokens are then partitioned into $B$ logical buckets $\{\mathcal{G}_b\}_{b=1}^{B}$ according to the smoothed mass, such that the total mass of each bucket is approximately balanced:
\begingroup

\begin{equation*}
\sum_{\omega\in \mathcal{G}_b} p(\omega)^{\alpha}
\approx
\frac{1}{B}\sum_{\omega\in\mathcal{V}\setminus \mathcal{V}_{\mathrm{vip}}} p(\omega)^{\alpha}.
\end{equation*}
\endgroup

In the main experiments, we use the following default settings:
\begingroup

\begin{equation*}
K_{\mathrm{vip}}=200,\qquad B=32,\qquad \rho=0.5,\qquad H=2,\qquad \alpha=0.5.
\end{equation*}
\endgroup
Here, $K_{\mathrm{vip}}$ denotes the size of the head set, $B$ denotes the number of logical buckets, $\rho=S/|\mathcal{V}|$ is the compression ratio, $H$ denotes the number of intra-bucket local hash paths used in dense buckets, and $\alpha$ is the smoothing exponent. In addition, each token is allowed to have at most three additional paths, and the additional paths are assigned geometrically decayed weights with decay factor $0.8$. These settings are used throughout the main experiments.

After the logical grouping is completed, we map tokens to concrete physical rows. For VIP tokens, we assign base rows in the VIP region through deterministic offsets; when needed, a small number of additional rows can also be assigned and combined with the base row through predefined weights. For ordinary tokens, we first determine the available physical subregion according to the assigned logical bucket, and then perform the intra-bucket mapping. If a bucket is sparse, tokens are mapped directly to physical rows according to their intra-bucket order, and the remaining unused rows can be recycled as alias rows. If a bucket is dense and collisions are unavoidable, we adopt multi-path local hashing within the bucket, generating a small number of intra-bucket retrieval paths for the same token and aggregating the corresponding results.

Accordingly, the retrieved representation of token $\omega$ at layer $\ell$ and branch $m$ can be further written as
\begingroup

\begin{equation*}
\mathbf{e}_{\ell}^{(m)}(\omega)
=
\frac{
\sum_{r=1}^{K_\omega}
\pi_{\omega,r}
\sum_{h=1}^{H}
\sigma\!\bigl(a_{\ell,\mathcal{M}(h,r,\omega)}^{(m)}\bigr)\,
\mathbf{B}_{\ell}^{(m)}[\mathcal{M}(h,r,\omega)]
}{
\sum_{r=1}^{K_\omega}\pi_{\omega,r}
}.
\end{equation*}
\endgroup

Here, $K_\omega$ denotes the number of valid candidate slots for token $\omega$, $\pi_{\omega,r}$ denotes the combination weight of the $r$-th candidate slot, $H$ denotes the number of local hash paths used within each candidate slot in the dense-bucket case, $a_{\ell,j}^{(m)}$ is a row-wise learnable parameter associated with physical row $j$, $\sigma(\cdot)$ denotes the Sigmoid activation function, and $\mathcal{M}(h,r,\omega)$ denotes the physical row index corresponding to the $h$-th local path under the $r$-th candidate slot for token $\omega$. The computation first aggregates the retrieved results along the $h$ dimension within each candidate slot, then weights them along the $r$ dimension by $\pi_{\omega,r}$, and finally normalizes by $\sum_{r=1}^{K_\omega}\pi_{\omega,r}$.

For VIP tokens and ordinary tokens in sparse buckets, the candidate slots mainly correspond to deterministically assigned base rows together with optional additional rows. For ordinary tokens in dense buckets, the candidate slots access a small number of physical rows through intra-bucket local hashing and aggregate the corresponding results. When additional paths are present, their combination weights follow a geometric decay rule: if a token is assigned the $c$-th additional path, the corresponding weight is given by a power of the decay factor, and is normalized together with the other candidate-slot weights during aggregation.

\subsection{Frequency-aware Routing (extended)}
\label{app:routing_knobs}
We summarize the key hyperparameters that govern the routing curve and the compression--trainability trade-off.
The compression ratio $\rho$ (equivalently, physical rows $S$) sets the global memory budget.
The smoothing exponent $\alpha$ controls how aggressively the long tail is softened when forming frequency-balanced buckets: smaller $\alpha$ makes the tail heavier and reduces extreme imbalances, while larger $\alpha$ pushes more mass into the head.
The number of buckets $B$ sets the granularity of mass balancing: larger $B$ yields finer balancing but smaller per-bucket capacity and potentially more fragmentation.
The alias size $K$ together with the decay profile over candidate slots controls how much unused capacity is recycled across buckets and provides robustness under high compression.
In implementation, dense buckets are realized with multi-path bucket-local hashing: a small number of bucket-local hash mappings with different multipliers generate retrieval paths for the same token, whose results are then aggregated.
The resulting access list $\mathcal{A}(\omega)$ is not necessarily a set, since different VIP aliases or bucket-local hash routes may revisit the same physical row.
This is why Eq.~(\ref{eq:alias_retrieval}) uses an effective aggregation coefficient $c_{\omega,j}$ together with the row-wise sigmoid gate $\sigma(a_{\ell,j}^{(m)})$, rather than interpreting each accessed row as having a uniquely normalized path-independent weight.
Together these knobs make the routing frequency-informed but not frequency-determined: we compute statistics once at negligible cost, and allow optimization to adjust effective allocation through alias mixing and gates when the raw frequency proxy is imperfect.

\subsection{Empirical Observation}\label{appobs:metrics}
We define the metrics used in Section~\ref{sec:observation}.
For a token $x_t$ at layer $\ell$, the injected vector is $\Delta_{\ell,t}$ as in Eq.~(\ref{eq:obs_inj}).
We measure \textbf{activation magnitude} for token type $\omega$ by
\begin{equation}
\mathcal{M}(\omega)=\frac{1}{|\mathcal{T}_\omega|}\sum_{t\in\mathcal{T}_\omega}\frac{1}{L}\sum_{\ell=1}^{L}\left\|\Delta_{\ell,t}\right\|_2,
\label{eq:obs_norm}
\end{equation}
where $\mathcal{T}_\omega$ is the set of token positions with $x_t=\omega$.

To quantify semantic drift, we compute the \textbf{cosine similarity} between a layer output $\mathbf{h}^{(c)}_{\ell,t}$ from component $c\in\{\text{attn},\text{ffn}\}$ and the input embedding $\mathbf{e}_{x_t}$:
\begin{equation}
\mathrm{Sim}^{(c)}_\ell=\mathbb{E}_{t}\left[\frac{\langle \mathbf{h}^{(c)}_{\ell,t}, \mathbf{e}_{x_t}\rangle}{\|\mathbf{h}^{(c)}_{\ell,t}\|_2\,\|\mathbf{e}_{x_t}\|_2}\right].
\label{eq:obs_sim}
\end{equation}

For redundancy across parallel slots, we report the pairwise cosine similarity between slot-level retrieved embeddings, averaged over sampled tokens, using the same formulation as Eq.~(\ref{eq:obs_sim}) but computed between slot pairs rather than against input embeddings.

\subsection{Stability rationale}\label{app:stability}
We provide brief justifications for the scalings introduced in Section~\ref{sec:method:opt}.

\paragraph{Why $1/\sqrt{M_\ell}$ in Eq.~(\ref{eq:triop}).}
Let $\mathbf{z}_m\in\mathbb{R}^{d_s}$ denote the injected vector from view $m$ for a fixed token position, after the per-view gate is applied.
Assume $\mathbb{E}[\mathbf{z}_m]=\mathbf{0}$ and $\mathrm{Cov}(\mathbf{z}_m)=\sigma^2\mathbf{I}$, and that different views are weakly correlated due to independent tables/extractors.
Then the covariance of the summed injection satisfies
\begin{equation}
\mathrm{Cov}\!\left(\frac{1}{\sqrt{M_\ell}}\sum_{m=1}^{M_\ell}\mathbf{z}_m\right)
=\frac{1}{M_\ell}\sum_{m=1}^{M_\ell}\mathrm{Cov}(\mathbf{z}_m)
\approx \sigma^2\mathbf{I},
\end{equation}
so the injected residual has a scale that is approximately invariant to the number of views.
Without this normalization, the residual magnitude grows with $M_\ell$, which can destabilize training when $M_\ell$ is increased.

\textbf{Why depth scaling uses $\sqrt{\ell+1}$ in Eq.~(\ref{eq:gate_rule}).}
Consider a simplified residual-stream model $\mathbf{H}_{\ell}=\mathbf{H}_{\ell-1}+\mathbf{r}_{\ell}$, where $\mathbf{r}_{\ell}$ is the block update with $\mathbb{E}[\mathbf{r}_{\ell}]=\mathbf{0}$ and $\mathrm{Var}(\mathbf{r}_{\ell})\approx \sigma_r^2$ (approximately constant across depth under common initializations).
Ignoring cross-layer correlations, we have $\mathrm{Var}(\mathbf{H}_{\ell})\approx (\ell+1)\sigma_r^2$, hence the RMS scale of the residual stream grows like $\sqrt{\ell+1}$\citep{xiong2020layer}.
Multiplying the injection gate by $\sqrt{\ell+1}$ keeps the injected residual at a comparable \emph{relative} scale across depth rather than shrinking in deeper layers.
The warmup factor $w_{\text{warmup}}(u)$ further prevents early optimization from being dominated by a sparsely trained injection channel.

\paragraph{Why we use a dimension-aware and sparse-aware step size.}
For an injected vector of dimensionality $d_s$, if the per-coordinate gradient variance is roughly constant, then $\mathbb{E}\| \nabla \|_2^2$ scales linearly with $d_s$.
To keep the expected update magnitude comparable when changing the injection dimensionality across sites (e.g., $d_s=d_{\text{kv}}$ for value injection versus $d_s=d$ for inter-layer injection), a natural rule is to scale the learning rate as $\eta_{\text{inj}}\propto 1/\sqrt{d_s}$.
In addition, lookup-table rows are updated sparsely: a row corresponding to a rare token receives far fewer updates, so under a uniform step size its cumulative parameter movement is disproportionately small.
Using a dedicated parameter group with a larger effective step size provides a coarse but effective correction for this sparsity mismatch, improving utilization of rarely visited rows under a fixed memory budget.

\paragraph{How the lookup learning-rate multiplier relates to table/vocabulary size.}
We make the dependence explicit with a simple update-frequency model.
Let $N$ be the total number of token positions processed so far (across all sequences), and let $q_j$ denote the probability that a position hits physical row $j$ in a given view (this depends on the routing, hashing, and aliasing).
Then the expected number of updates to row $j$ is $n_j \approx N q_j$.
If the per-update gradient statistics are roughly stationary, then the cumulative parameter movement of row $j$ behaves like a random walk and scales as
\begin{equation}
\mathbb{E}\|\Delta \mathbf{B}[j]\|_2 \;\propto\; \eta_j \sqrt{n_j}\;=\;\eta_j\sqrt{N q_j}.
\label{eq:rw_update}
\end{equation}
To keep rarely visited rows from lagging behind purely due to update scarcity, a natural sparse-aware rule is
\begin{equation}
\eta_j \propto \frac{1}{\sqrt{q_j}},
\label{eq:lr_q}
\end{equation}
which makes $\eta_j\sqrt{q_j}$ approximately constant across rows.

This immediately yields the vocabulary/table-size scaling in the uniform-hit regime.
If routing is close to uniform over a table of $S$ rows, then $q_j\approx 1/S$ and Eq.~(\ref{eq:lr_q}) suggests a multiplier that grows like
\begin{equation}
\eta_{\text{lookup}} \propto \sqrt{S}.
\label{eq:lr_sqrtS}
\end{equation}
Since $S=\rho|\mathcal{V}|$ under compression ratio $\rho$, this implies $\eta_{\text{lookup}} \propto \sqrt{\rho|\mathcal{V}|}$ as a coarse baseline.

In practice, $q_j$ is far from uniform: VIP rows have much larger hit rates than hashed body rows, and aliasing spreads probability mass across multiple candidates.
We therefore implement Eq.~(\ref{eq:lr_q}) at the granularity of parameter groups rather than per-row: we use a larger multiplier for hashed/aliased rows than for VIP rows, and cap the multiplier to prevent instability on the head.
Finally, we apply the same warmup factor $w_{\text{warmup}}(u)$ to the lookup learning-rate multiplier so that the injection channel does not dominate optimization before the memory and gates have started to learn.

\subsection{Why linear stacking does not increase capacity}\label{appobs:linear}
We sketch a simple argument for why adding multiple linear parameters to the same $n$-gram slot does not increase effective capacity without a nonlinear, view-specific extractor.
Let $z(x)\in\mathbb{R}^d$ be the retrieved vector for an $n$-gram slot, and consider a model that adds $K$ linear transforms of $z$ to the same stream:
\begin{equation}
\sum_{k=1}^{K} \mathbf{W}_k z(x) = \mathbf{W}_{\mathrm{eq}} z(x), \quad \mathbf{W}_{\mathrm{eq}} \triangleq \sum_{k=1}^{K} \mathbf{W}_k.
\end{equation}
Thus, the composition is equivalent to a single linear map, and the representational subspace spanned by the slot remains unchanged.
To increase capacity, one needs either nonlinear mixing or view-specific transformations that produce distinct features before fusion; this motivates the use of normalized gated ShortConv in Section~\ref{sec:method:shortconv}.

\subsection{Why ShortConv increases expressivity}\label{app:shortconv_rationale}
We briefly sketch why a view-specific, gated local convolution expands the set of representable features compared to context-independent lookup and linear stacking.

\paragraph{From token identity to $x$-gram features.}
Plain lookup produces a context-independent feature: at position $t$, the retrieved vector depends only on $x_t$.
By contrast, a causal convolution of width $k$ produces features that depend on a window $(x_{t-k+1},\ldots,x_t)$, i.e., an $x$-gram of length up to $k$.
This strictly enlarges the function class: even with fixed lookup rows, convolving the retrieved sequence can represent order- and proximity-sensitive patterns (e.g., ``not good'' vs.\ ``good'') that cannot be expressed by adding token-wise vectors.

\paragraph{Why gating and multi-scale matter.}
The multiplicative gate makes the extractor input-dependent: it can suppress a feature when it is spurious in the current context (e.g., due to hash collisions) and amplify it when it aligns with nearby cues.
Using multiple kernel sizes further increases coverage of local phenomena at different spans and provides multiple, complementary views of the same memory rows.
Since each view $m$ uses its own kernels, the extractor also breaks symmetry across parallel tables, which mitigates collapse to redundant subspaces before fusion.

\paragraph{Why residual fusion is critical.}
Residual fusion $\tilde{\mathbf{E}}=\mathbf{E}+\mathbf{C}$ serves two roles.
First, it preserves a stable identity anchor: even if the convolutional branch is noisy early in training, the model can fall back to the raw retrieval and learn to use extracted features gradually.
Second, under sparse updates, it maintains a direct gradient path to the memory rows regardless of gate saturation, improving trainability of rarely visited rows.
Together, these properties explain why ShortConv tends to improve effective diversity and scaling behavior beyond what can be achieved by simply summing additional embeddings.

\clearpage

\end{document}